\documentclass[sigconf]{acmart}
 
\AtBeginDocument{%
  }

\setcopyright{acmlicensed}
\copyrightyear{2026}
\acmYear{2026}
\acmDOI{XXXXXXX.XXXXXXX}
\acmConference[Conference acronym 'XX]{Make sure to enter the correct
  conference title from your rights confirmation email}{June 03--05,
  2018}{Woodstock, NY}
\acmISBN{978-1-4503-XXXX-X/2018/06}
\renewcommand\footnotetextcopyrightpermission[1]{}%remove DOI information part
\acmSubmissionID{4751}
\usepackage{amsmath}
\usepackage{booktabs}   % 优化表格线
\usepackage{multirow}   % 合并单元格
\usepackage{amssymb}    % 提供 \checkmark
\usepackage{pifont}     % 提供 \xmark
\usepackage[table]{xcolor} % 提供表格颜色支持
\begin{document}

\title{VistaRef: Boosting Visual Spatial Orientation Awareness for Pointing-to-Object Detection}
\thanks{\textbf{Code:\url{https://github.com/lingli1724/VistaRef}}}
\author{Ling Li}
\affiliation{%
  \institution{Tsinghua University}
  \state{Beijing Shi}
  \country{China}}
\email{liling25@mails.tsinghua.edu.cn}

\author{Zhizhen Cai}
\affiliation{%
  \institution{Beijing University of Posts and Telecommunications}
  \state{Beijing Shi}
  \country{China}}
\email{Chua_ZH2@outlook.com}

\author{Xinkun Wu}
\affiliation{%
  \institution{Tsinghua Shenzhen International Graduate School}
  \city{Shenzhen}
  \state{Guangdong}
  \country{China}}
\email{wuxk25@mails.tsinghua.edu.cn}

\author{Ziyu Zhu}
\affiliation{%
  \institution{Tsinghua University}
  \state{Beijing Shi}
  \country{China}}
\email{zhuziyu.edward@gmail.com}

\author{Jiaqing Lyu}
\affiliation{%
  \institution{Tsinghua University}
  \state{Beijing}
  \country{China}}
\email{lyujq25@mails.tsinghua.edu.cn}

\author{Bowen Liu}
\affiliation{%
  \institution{Dalian University of Technology}
  \state{Dalian Shi}
  \country{China}}
\email{liubw20050702@mail.dlut.edu.cn}

\author{Zhidong Deng}
\affiliation{%
  \institution{Dalian University of Technology}
  \state{Dalian Shi}
  \country{China}}
\email{michael@tsinghua.edu.cn}
\renewcommand{\shortauthors}{Trovato et al.}

%%%%%%%%%%%%%%%%%%%%%%%%%%%%%%%%%%%%%%%%%%%%%%%%%%%%%%%%%%   摘要
\begin{abstract}
  Grounding deictic gestures in natural images is fundamental to AR and human-robot collaboration, providing a basis for seamless spatial interaction. While Transformer-based visual models have achieved significant progress in general object detection, their global attention mechanisms often neglect micro-geometric relationships, degrading orientation accuracy. In pointing tasks, this deficiency manifests as an inability to accurately capture the pointing ray implied by finger poses, which results in pointing drift and localization ambiguity when dealing with distant or densely packed objects. To address this, we propose \textbf{VistaRef}, a framework designed to explicitly enhance spatial orientation awareness. First, we develop the Local Hand Entity Modeling (LHEM) module, which incorporates hand-pose embeddings to strengthen the model’s capability to capture subtle finger deviations. Second, drawing inspiration from multi-view geometry, we construct the Geometric Ray Modeling (GRM) module to transform implicit orientation information into explicit spatial geometric features, guiding feature aggregation and deep fusion via attention mechanisms. Furthermore, we introduce a novel Orientation-Consistent Alignment Loss (OCAL) to synergistically supervise hand presence and pointing consistency, ensuring that all architectural improvements collectively serve the core objective of spatial localization. 
  Experimental results demonstrate that VistaRef significantly outperforms the baseline, achieving a \textbf{14}-point absolute gain in grounding accuracy. Qualitative analysis further confirms that VistaRef effectively models the geometric correlation from hand to target, bridging the spatial perception gap inherent in traditional Transformers for complex scenarios. Code: \url{https://github.com/lingli1724/VistaRef.}
\end{abstract}

%%%%%%%%%%%%%%%%%%%%%%%%figure%%%%%%%%%%%%%%%%%%%%%%%%%%

%%%%%%%%%%%%%%%%%%%%%%%%%%%%%%%%%%%%%%%%%%%%%%%%%%%%%%%%%%%%%%%
\received{20 February 2007}
\received[revised]{12 March 2009}
\received[accepted]{5 June 2009}

\maketitle

\section{Introduction}
%%%%%%%%%%%%%%%%%%%%%%%%%%%%%%%%figure1%%%%%%%%%%%%%%%%%%%%%%%%%%%%%%%%%
\begin{figure}[b] % 注意：这里绝对不能有星号
    \centering
    % 使用 1.0 倍的单栏宽度
    \includegraphics[width=\columnwidth]{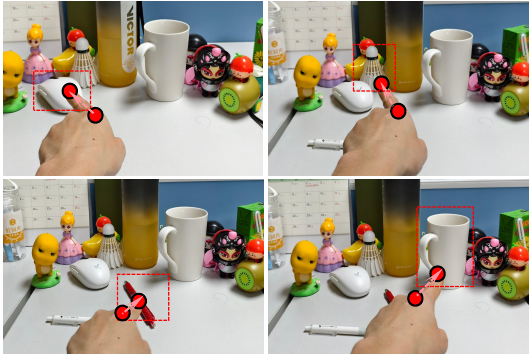} 
    \caption{Examples of pointing-to-object detection. The target object varies relative to the finger's orientation; even subtle shifts in pointing direction lead to different targets, necessitating robust spatial awareness from the model.}
    \label{fig:pointing_variation}
\end{figure}
%%%%%%%%%%%%%%%%%%%%%%%%%%%%%%%%%%%%%%%%%%%%%%%%%%%%%%%%%%%%%%%%%%%%%%%%
%%%%%%%%%%%%%%%%%%%%%%%%%%%%figure2%%%%%%%%%%%%%%%%%%%%%%%%%%%%%%%%%%%%%  图2

\begin{figure*}[t]
    \centering
    \includegraphics[width=\linewidth]{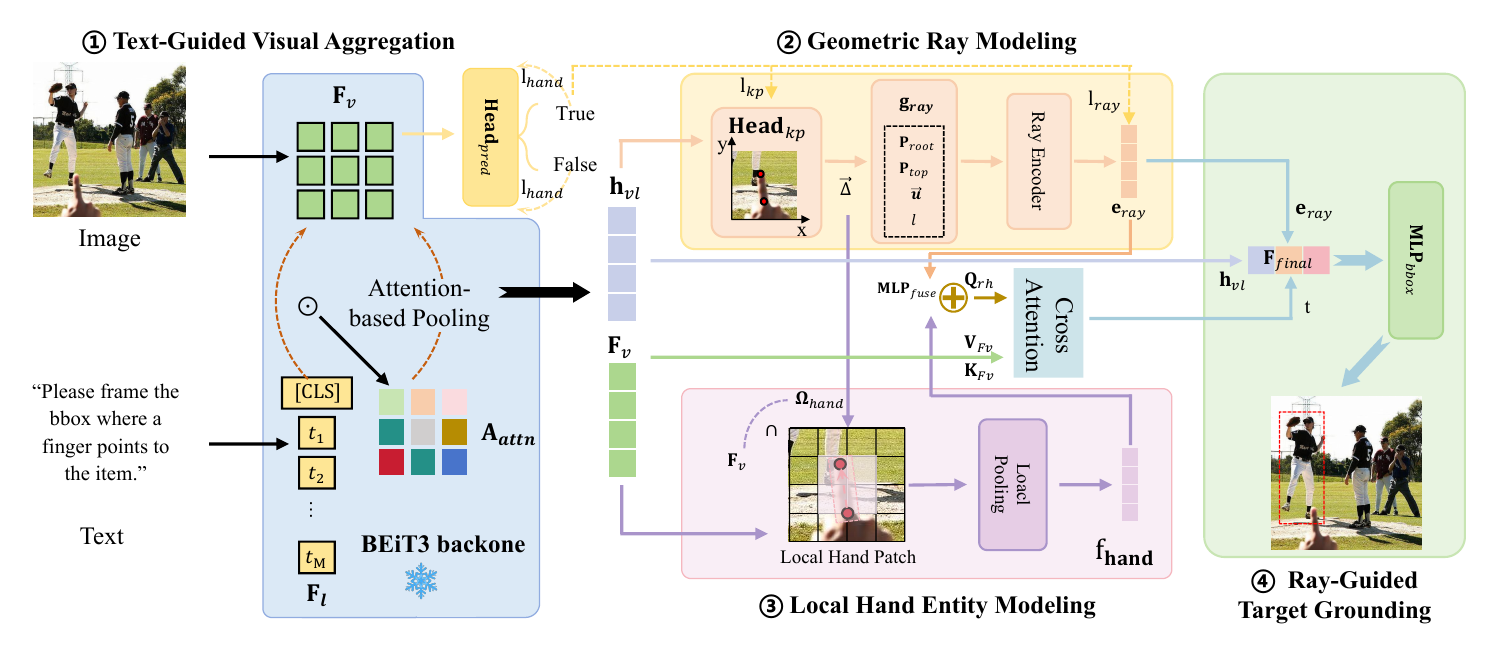}
    \caption{\textbf{Overview of the VistaRef framework}. Our model consists of four hierarchical modules: \textbf{(1) Text-Guided Visual Aggregation} extracts target-aware semantic representations via asymmetric masking; \textbf{(2) Geometric Ray Modeling (GRM)} regresses directional primitives to encapsulate hand pointing orientation; \textbf{(3) Local Hand Entity Modeling (LHEM)} purifies gestural features from a dynamically proposed region $\Omega_{hand}$; \textbf{(4) Ray-Aware Target Grounding} performs precise object localization through ray-guided interaction modeling.}
    \label{fig:vista_ref_overview}
    \vspace{-10pt}
\end{figure*}

%%%%%%%%%%%%%%%%%%%%%%%%%%%%%%%%%%%%%%%%%%%%%%%%%%%%%%%%%%%%%%%%%%%%%
Within the workflow of Visual Grounding (VG)  \cite{deng2022transvgendtoendvisualgrounding,pantazopoulos2025understandingvisualgroundingvisual,he2023improvedvisualgroundingselfconsistent}, Referring Expression Comprehension (REC)   \cite{9470913,qiao2020referringexpressioncomprehensionsurvey,goto2025referringexpressioncomprehensionsmall,yu2018mattnetmodularattentionnetwork} and Referring Expression Segmentation (RES)  \cite{liu2023gresgeneralizedreferringexpression,wu2023advancingreferringexpressionsegmentation,wu2024rgsanruleguidedspatialawareness} establish the correspondence between linguistic semantics and visual entities through the prediction of bounding boxes and instance masks, respectively, serving as the logical foundation for agent environment perception. However, conventional methods are often constrained by the inherent scarcity of referential information in text-only queries, failing to accommodate the physical deictic cues of interlocutors in real-world spaces. For example, in scenarios containing multiple objects with similar characteristics, agents struggle to accurately resolve the user's referential intent without deictic cues (e.g., body language) \cite{li2022understandingembodiedreferencetouchline}. To transcend these limitations, Embodied Reference Understanding (ERU)  \cite{lu2024scaneru,shi2022spatial,guo2025objectcategoriesmultiattributereference,nakamura2023deepointvisualpointingrecognition,chen2021yourefitembodiedreferenceunderstanding,mane2025ges3vigincorporatingpointinggestures,islam2025embodiedreferringexpressioncomprehension,nakamura2023deepointvisualpointingrecognition} requires models to interpret physical intentions through spatial perception such as pointing gestures \cite{li2022understandingembodiedreferencetouchline,guo2024addinoattentiondynamicdinodistanceaware}, body orientation \cite{nakamura2023deepointvisualpointingrecognition}, and gaze projection  \cite{johari2021gaze,qian2023gvgnet}, emphasizing the embodied nature of interaction.

Under this framework, pointing-to-object detection serves as a key sub-task \cite{li2026languagegroundingreferringexpressions,guo2024addinoattentiondynamicdinodistanceaware,chen2021yourefitembodiedreferenceunderstanding}. This task aims to overcome the limitations of ambiguous semantic alignment by explicitly formulating a hierarchical physical modeling chain, progressing from body keypoint estimation to pointing vector construction and culminating in target object projection, thereby achieving deterministic ray alignment, as shown in Fig. \ref{fig:pointing_variation}. By empowering models with precise physical spatial awareness for high-frequency collaboration in cluttered environments, pointing-to-object detection plays a vital role in the fields of augmented reality (AR) \cite{AR1,reg,OSTAR,kao2022designing}, human-robot collaboration \cite{Liu_2024,zhao2026,paralikar2025robot,10.1145/3382507.3418863}, and human–computer interface \cite{yi2022magnetic,lv2022deep,xu2022robust,gupta2023survey}.

% Pointing-to-object detection is of paramount importance in the realms of Augmented Reality (AR) \cite{AR1,reg,OST-AR}, human-robot collaboration \cite{Liu_2024,zhao2026,paralikar2025robot,10.1145/3382507.3418863}, and natural Human-Computer Interaction (HCI) \cite{yi2022magnetic,lv2022deep}. This task requires systems to accurately decipher human interactive intent by transforming the directional semantics implicit in hand gestures into precise spatial localizations of specific objects. Unlike conventional Visual Grounding \cite{yu2018mattnetmodularattentionnetwork,deng2022transvgendtoendvisualgrounding,liu2024groundingdinomarryingdino,dai2024simvgsimpleframeworkvisual}, pointing-based tasks do not rely solely on the visual features of the target; rather, they are deeply coupled with the geometric projection relationship \cite{li2022understandingembodiedreferencetouchline,guo2025objectcategoriesmultiattributereference,guo2025objectcategoriesmultiattributereference} between the hand pose and the target object.

In recent years, single-stage Transformer-based frameworks \cite{liu2024groundingdinomarryingdino,li2022groundedlanguageimagepretraining,liu2024groundingdinomarryingdino,xiao2024onerefunifiedonetowerexpression} have significantly enhanced the efficiency of VG through end-to-end attention mechanisms. However, applying such models to pointing-to-object detection tasks remains challenging, as the global attention mechanism of Transformers tends to prioritize coarse-grained semantic associations at the expense of fine-grained micro-scale spatial geometric constraints  \cite{chu2023conditionalpositionalencodingsvision,kim2023understandinggaussianattentionbias,dong2022cswintransformergeneralvision}. This leads to a significant orientation insensitivity, where the model fails to generate corresponding variations in physical features in response to subtle deviations in finger pose \cite{sundermeyer2019implicit3dorientationlearning,tong2026mvhoibridgemultiviewcondition,bertens2025symmetry}. In complex scenarios involving distant targets or cluttered environments, this lack of spatial awareness directly leads to severe localization drift and predictive ambiguity  \cite{li2022understandingembodiedreferencetouchline,guo2024addinoattentiondynamicdinodistanceaware}.

We contend that the key to resolving pointing ambiguity lies in transforming implicit geometric priors into explicit feature constraints. A successful pointing system must possess two core capabilities: high-precision pose perception of the pointing origin (i.e., the hand entity)  \cite{cao2019openposerealtimemultiperson2d,zhou2024simplebaselineefficienthand}and robust modeling of the geometric path from the hand to the target \cite{li2022understandingembodiedreferencetouchline,guo2024addinoattentiondynamicdinodistanceaware}. To this end, we aim to introduce a spatial geometric alignment mechanism upon the model of OneRef \cite{xiao2024onerefunifiedonetowerexpression}. This allows the model to perform feature aggregation along a deterministic geometric ray—much like the human visual system—thereby overcoming the inherent limitations of pure attention mechanisms in spatial reasoning.

Based on these insights, we propose the VistaRef framework. To accurately capture the pointing ray, we introduce the Local Hand Entity Modeling (LHEM) module, which strengthens the perception of subtle fingertip movements through hand-pose embeddings. For pointing-path modeling, we design the Geometric Ray Modeling (GRM) module, which utilizes attention mechanisms to deeply fuse orientation information with visual features, explicitly guiding the model to focus on the spatial regions covered by the ray. Furthermore, to ensure synergy between modules, we propose the Orientation-Consistent Alignment Loss (OCAL), which employs geometric supervision to force the model to learn physically self-consistent spatial mappings. The core advantage of VistaRef lies in its dual focus: it considers not only "where the object is" but also "how the hand points to it," fundamentally enhancing the robustness of spatial perception.

Our main contributions are summarized as follows:

% (1) We unveil the limitations of traditional Transformers regarding insufficient spatial spatial orientation awareness in pointing tasks and emphasize the necessity of pointing ray modeling.

% (2) We propose the VistaRef framework, which achieves hierarchical geometric modeling—from local gestures to global spatial rays—via the Local Hand Entity Modeling and GRM modules.

% (3) We design the Orientation-Consistent Alignment Loss, introducing explicit geometric alignment constraints into end-to-end Transformer learning.

% (4) Experimental results demonstrate that VistaRef significantly outperforms mainstream models, exhibiting superior localization robustness, particularly in complex pointing scenarios.
%%%%%%%%%%%%%%%%ling's version%%%%%%%%%%%%%%%%%%%
(1) \textbf{Identification of Orientation Insensitivity:} We reveal the inherent spatial orientation insensitivity of traditional Transformers in pointing tasks and emphasize the foundational role of pointing ray modeling.

(2) \textbf{The VistaRef Framework:} We propose VistaRef, a novel framework that implements a hierarchical geometric modeling strategy—bridging micro-level gestural cues with macro-level spatial trajectories—enabled by the LHEM and GRM modules.

(3) \textbf{Orientation-Consistent Supervision:} We introduce the OCAL loss function, which incorporates explicit geometric constraints into the end-to-end Transformer learning process to ensure physically self-consistent spatial alignment.

(4) \textbf{Systematic Validation:} Extensive experiments demonstrate that VistaRef significantly outperforms state-of-the-art baselines, exhibiting superior localization robustness and orientation accuracy, particularly in cluttered and distant pointing scenarios.

%%%%%%%%%%%%%%%%%%%%%%%%%figure3%%%%%%%%%%%%%%%%%%%%%%%%%%%%%%%%%%%%%%%%  图3
\begin{figure*}[t]
    \centering
    \includegraphics[width=\linewidth]{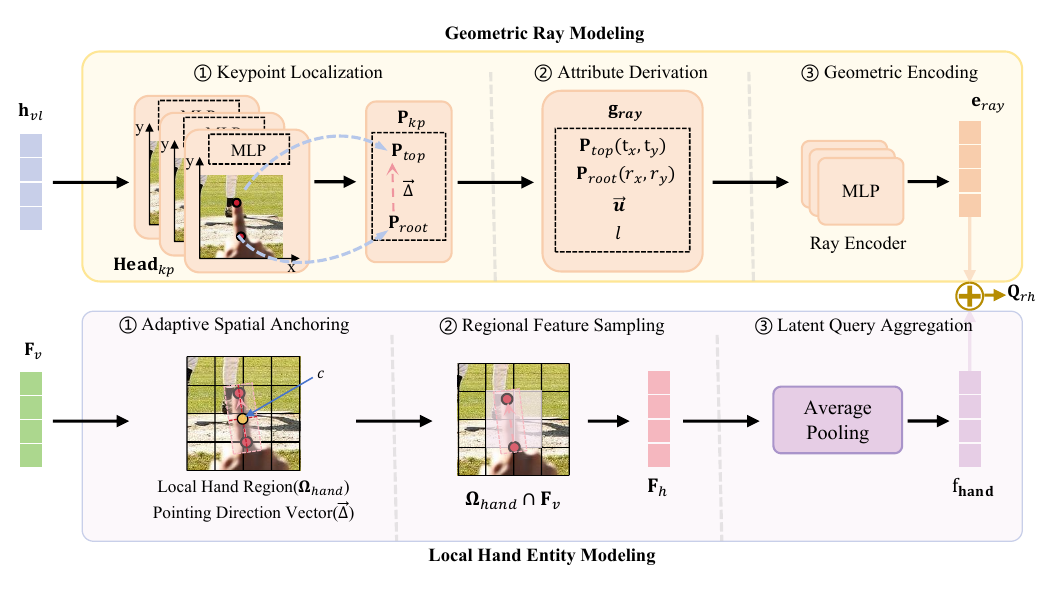}
    \caption{\textbf{Illustration of the Geometric Ray Modeling (GRM) and Local Hand Entity Modeling (LHEM).}}
    \label{fig:ray_hand_modeling}
\end{figure*}
%%%%%%%%%%%%%%%%%%%%%%%figure%%%%%%%%%%%%%%%%%%%%%%%%%%%%%%%%%%%%%%%%%

%%%%%%%%%%%%%%%%%%%%%%%%%%%%%%%%%%%%%%%%%%%%%%%%%%%%%%%%%%%%%%%%%%%%%%%%%%%%%%%%%%%%%%%%%%%%%%%%%%%%%%%%%%%%%%%%%%%%%%%%%%%%%%%%%%%%%%%%%%%%%%%%
\vspace{-10pt}
\section{Related Work}
%%%%%%%%%%%%%%%%%%%%%%%%%%%%%%%%%%%%%%%%%%%%%%%%%%   2.1  % 视觉定位与指称表达理解
\subsection{Visual Grounding and Referring Expression Comprehension} % 视觉定位与指称表达理解

In the realm of VG \cite{Xiao_2026,pantazopoulos2025understandingvisualgroundingvisual,wang2024learningvisualgroundinggenerative}, REC \cite{qiao2020referringexpressioncomprehensionsurvey,zhao2024rethinking,han2024zero,lu2024lgr} focuses on mapping complex linguistic references onto spatial bounding boxes in images.
Early two-stage frameworks, such as MAttNet \cite{yu2018mattnetmodularattentionnetwork} and NMTREE \cite{liu2019learningassembleneuralmodule}, decomposed expressions into modular components or dependency trees to achieve interpretable reasoning. To enhance computational efficiency, one-stage models like RealGIN \cite{9470913} and TransVG \cite{deng2022transvgendtoendvisualgrounding} shifted the workflow toward direct coordinate regression.
Following this trend, subsequent research has further deepened the multimodal modeling capabilities of Transformers, pursuing higher-dimensional task unification  \cite{Zhu_2022,Xiao_2024,dai2025propvgendtoendproposaldrivenvisual,he2023grecgeneralizedreferringexpression} and deep modal alignment \cite{liu2024groundingdinomarryingdino,xiao2024onerefunifiedonetowerexpression}. Task unification not only covers the integration of multiple referring tasks but also includes adaptation to referring requirements at different granularities, while deep alignment focuses on achieving more efficient synchronization between textual and visual semantics, thereby significantly improving localization accuracy. Specifically, the unified network proposed by SeqTR completes phrase grounding, REC, and RES in a one-stop manner via discrete coordinate point sequence prediction \cite{Zhu_2022}. Additionally, HiVG  \cite{Xiao_2024} and PropVG  \cite{dai2025propvgendtoendproposaldrivenvisual}introduce hierarchical or multi-granularity modeling mechanisms, significantly enhancing the dimension of perception. GREC \cite{he2023grecgeneralizedreferringexpression} addresses the limitations of traditional REC, which only supports single targets, by providing generalized support for multi-target and no-target expressions.
As a typical representative of multimodal deep alignment, Grounding DINO \cite{liu2024groundingdinomarryingdino} implements full-stage cross-modal fusion. To deepen modal integration, OneRef \cite{xiao2024onerefunifiedonetowerexpression} adopts a single-tower architecture coupled with Masked Referring Expression Modeling to capture complex referential relationships.
Despite the progress made by previous work in high-dimensional task unification and semantic alignment, such models still treat coordinates as static indices, failing to perceive physical space and human pointing intentions.
\vspace{-10pt}
%%%%%%%%%%%%%%%%%%%%%%%%%%%%%%%%%%%%%%%%%%%%%%%%%%   2.2  % 具身指代理解

\subsection{Embodied Reference Understanding and Hand Pointing Detection}%讨论手物之间的交互研究， 手指指向检测
Human communication is multimodal and embodied, with users frequently leveraging multimodal deictic cues to convey intent \cite{johnson2015embodied,eyiokur2026cape,guo2024addinoattentiondynamicdinodistanceaware,mane2025ges3vigincorporatingpointinggestures,nakamura2023deepointvisualpointingrecognition}. This necessitates a transition from heuristic semantic alignment toward task-specific geometric modeling within the scope of ERU \cite{lu2024scaneru,shi2022spatial,guo2025objectcategoriesmultiattributereference,nakamura2023deepointvisualpointingrecognition,chen2021yourefitembodiedreferenceunderstanding,mane2025ges3vigincorporatingpointinggestures,islam2025embodiedreferringexpressioncomprehension,nakamura2023deepointvisualpointingrecognition}. To address spatial perspective-taking and parallax, REP  \cite{shi2023spatialvisualperspectivetakingview} employs view rotation to simulate the sender's egocentric perception. To further refine pointing precision, AD-DINO  \cite{guo2024addinoattentiondynamicdinodistanceaware} integrates distance-aware mechanisms with dynamic attention source prediction. Touch-Line Transformer \cite{li2022understandingembodiedreferencetouchline} introduces the virtual touch line, utilizing geometric consistency loss to mathematically constrain the gesture-target alignment. In the 3D domain, DeePoint \cite{nakamura2023deepointvisualpointingrecognition}  attempts to jointly model holistic spatio-temporal features for end-to-end pointing recognition and 3D direction estimation. However, the model's 2D pose estimator, OpenPifPaf \cite{kreiss2021openpifpafcompositefieldssemantic}, needs to detect 17 human joint keypoints per frame. This undoubtedly escalates prediction complexity, while various body keypoints remain susceptible to occlusions.

To resolve these problems, we propose VistaRef, which explicitly formulates a hierarchical physical modeling process progressing from hand reference points through pointing vectors to spatial rays. This geometric-driven workflow significantly streamlines the keypoint prediction pipeline while accurately capturing the underlying physical mechanics of pointing, thereby facilitating robust ray alignment in cluttered scenarios.

\vspace{-10pt}
%%%%%%%%%%%%%%%%%%%%%%%%%%%%%%%%%  3、方法
\section{Method}

In this section, we present \textbf{VistaRef}, a task-specific framework engineered to mitigate referential ambiguity by explicitly modeling the \textbf{``point-to-line''} physical progression of pointing gestures. VistaRef maps abstract multimodal intent onto deterministic geometric rays, which serve as structural priors to guide fine-grained VG. As illustrated in Fig.~\ref{fig:vista_ref_overview}, given a linguistic query such as \textit{``Please frame the bbox where a finger points to the item''} the model regresses the coordinates of the hand's root and tip, extracts distilled gestural features from the localized hand region, and effectively traverses the synthesized ray to precisely localize the target entity. As show in Fig.~\ref{fig:vista_ref_overview}, VistaRef adopts a cascaded architecture consisting of \textbf{Text-Guided Visual Aggregation} (Sec. \ref{sec:Cross-modal Feature Representation}), \textbf{Geometric Ray Modeling (GRM)} (Sec. \ref{sec:Geometric Ray Modeling}), \textbf{Local Hand Entity Modeling (LHEM)} (Sec. \ref{sec:Local Hand Entity Modeling}), and \textbf{Ray-Aware Target Grounding} (Sec. \ref{sec:Ray-guided Interaction Modeling and Box Regression}).

For each image-text pair, we utilize a BEiT3-based backbone to encode the joint vision-language input into a target-aware semantic representation $\mathbf{h}_{vl}$. Building upon this context, the GRM module regresses hand coordinates to formulate a directional prior $\mathbf{e}_{\text{ray}}$, while the LHEM module extracts purified gestural features $\mathbf{f}_{\text{hand}}$ from a dynamically proposed local region $\Omega_{\text{hand}}$ (Fig.~\ref{fig:ray_hand_modeling}). These components are synthesized into a ray-aware hand query $\mathbf{Q}_{\text{rh}}$, which interacts with global visual features via a cross-attention to capture the spatial associations between the hand and the potential target. Subsequently, the resulting interaction features $\mathbf{t}$ are non-linearly fused with the global context to predict the final object bounding box. Finally, the framework is optimized via a \textbf{Conditional Supervision Strategy} (Sec. \ref{sec:Training Objectives and Conditional Supervision}), which gates regression branches based on gesture presence to ensure robust geometric alignment.

\vspace{-10pt}
%%%%%%%%%%%%%%  3.1、跨模态特征提取/表达-交互
\subsection{Cross-modal Feature Representation and Interaction}
\label{sec:Cross-modal Feature Representation}

We employ the highly versatile and unified BEiT-3 \cite{wang2022imageforeignlanguagebeit} architecture as our backbone. To fully leverage the robust and diverse multimodal representations acquired during the pre-training phase, its parameters are kept entirely frozen throughout the training process. The system directly processes raw images and text prompts, utilizing multimodal features extracted from the frozen backbone to provide precise semantic guidance for the subsequent spatial orientation awareness tasks.

The backbone initially encodes the input into a unified embedding space, yielding a visual feature sequence $\mathbf{F}_v$ and a linguistic sequence $\mathbf{F}_l$. To effectively capture cross-modal semantic consistency, we retain the original \textbf{Text-guided Visual Aggregation} mechanism. Specifically, a cross-modal affinity matrix $\mathbf{A}_{attn}$  is derived by computing the cosine similarity---implemented via dot-product interaction---between the global linguistic representation ([CLS] token) and discrete visual patches, where$\mathbf{A}_{attn} \in \mathbb{R}^{N}$, $N$ is the number of patches. Each entry in $\mathbf{A}_{attn}$ reflects the matching degree between the corresponding spatial patch and the linguistic query. Subsequently, we perform weighted information aggregation across visual patches based on $\mathbf{A}_{attn}$ to synthesize a holistic vision-language feature $\mathbf{h}_{vl}$, which encapsulates highly condensed target semantic information and serves as a robust foundation for subsequent fine-grained localization:
\begin{equation}
\mathbf{h}_{vl} = \sum_{i=1}^{N} \left( \sigma(\mathbf{A}_{attn, i}) \cdot \mathbf{F}_{v, i} \right)
\end{equation}
where $\sigma(\cdot)$ denotes the Softmax normalization operation, ensuring that the distribution weights of spatial attention sum to unity.

%%%%%%%%%%%%%%   3.2 几何射线建模
\subsection{Geometric Ray Modeling}
\label{sec:Geometric Ray Modeling}

To bridge the gap between abstract linguistic instructions and concrete spatial orientations, we propose a \textbf{GRM} module, as illustrated in Fig.~\ref{fig:ray_hand_modeling}. This module transforms global semantic context into an explicit directional representation---the ray---to capture the geometric essence of the pointing gesture.\\
\textbf{Keypoint Localization.} We employ a dedicated keypoint detection head $\mathbf{Head}_{kp}$, consisting of a three-layer Multilayer Perceptron (MLP) stack. Leveraging the text-guided holistic visual feature $\mathbf{h}_{vl}$, this head explicitly maps the normalized coordinates of the hand root $\mathbf{p}_r$ and the fingertip $\mathbf{p}_t$:
\begin{equation}
\{\mathbf{p}_r, \mathbf{p}_t\} = \sigma(MLP_{kp}(\mathbf{h}_{vl})), \quad \mathbf{p}_r, \mathbf{p}_t \in [0, 1]^2
\end{equation}
By utilizing $\mathbf{h}_{vl}$ as the input, the localized keypoints are inherently aligned with the linguistic query.\\
\textbf{Geometric Modeling.} 
Based on the predicted coordinates, we derive a set of explicit geometric attributes to characterize the pointing vector. Specifically, the magnitude $l = \|\Delta\|_2$ captures the gesture's scale, while the unit vector $\mathbf{u} = \Delta/l$ isolates its orientation.\\
\textbf{Geometric Encoding.} 
To project these low-dimensional geometric priors into a high-dimensional latent space, we concatenate them into a 7-dimensional geometric vector $\mathbf{g}_{ray} = [\mathbf{p}_r; \mathbf{p}_t; \mathbf{u}; l]$. Subsequently, a three-layer MLP serves as the Ray Encoder to transform $\mathbf{g}_{ray}$ into the ray embedding $\mathbf{e}_{ray} \in \mathbb{R}^d$:
\begin{equation}
\mathbf{e}_{ray} = MLP_{ray}(\mathbf{g}_{ray})
\end{equation}

%%%%%%%%%%%%%%%%%%%%%%%%%%%%%%%%%%%%%%%%%%%%%%%%%%表格1 总的实验结果表
%%%%%%%%%%%%%%%%%%%%%%%%%%%%%%%%%%%%%%%%%%%%%%%%%%%%%%%%%%%%%%%%%%%
\begin{table*}[t]
\centering
\caption{\textbf{Quantitative comparison with state-of-the-art methods in EgoPoint-Ground\cite{li2026languagegroundingreferringexpressions}.} We evaluate performance across three settings: Zero-shot, Train on Hybrid, and Train-Real Data Test. Precision at various IoU thresholds (P@0.3, 0.5, 0.7) and mean IoU (mIoU) are reported. The best results are highlighted in \textbf{bold}, and the second best (non-Ours methods) are \underline{underlined}.}
\label{tab:main_results}
\resizebox{\textwidth}{!}{%
\begin{tabular}{llcccc|cccc|cccc}
\toprule
\multirow{2}{*}{\textbf{Method}} & \multirow{2}{*}{\textbf{Backbone}} & \multicolumn{4}{c|}{\textbf{Zero-shot}} & \multicolumn{4}{c|}{\textbf{Train on Hybrid}} & \multicolumn{4}{c}{\textbf{Train-Real Data Test}} \\
\cmidrule(lr){3-6} \cmidrule(lr){7-10} \cmidrule(lr){11-14}
& & P@0.3 $\uparrow$ & P@0.5 $\uparrow$ & P@0.7 $\uparrow$ & mIoU $\uparrow$ & P@0.3 $\uparrow$ & P@0.5 $\uparrow$ & P@0.7 $\uparrow$ & mIoU $\uparrow$ & P@0.3 $\uparrow$ & P@0.5 $\uparrow$ & P@0.7 $\uparrow$ & mIoU $\uparrow$ \\
\midrule
\rowcolor[gray]{.95} \multicolumn{14}{l}{\textit{VLM-based methods}} \\
PropVG & BEiT-3-Base & 0.1469 & 0.0307 & 0.0044 & 0.1553 & 0.7471 & 0.6865 & 0.5785 & 0.6105 & 0.7095 & 0.6165 & 0.4822 & 0.5585 \\
PropVG & BEiT-3-Large & 0.1352 & 0.0249 & 0.0015 & 0.1470 & \underline{0.7653} & \underline{0.7191} & \underline{0.6299} & \underline{0.6441} & \underline{0.8414} & \underline{0.8244} & \underline{0.7848} & \underline{0.7745} \\
\midrule
\rowcolor[gray]{.95} \multicolumn{14}{l}{\textit{Grounding-based methods}} \\
HiVG & CLIP-B & \underline{0.3704} & 0.2616 & 0.1835 & 0.2929 & 0.7386 & 0.6442 & 0.4606 & 0.5458 & 0.8050 & 0.7031 & 0.4838 & 0.5794 \\
HiVG & CLIP-L & 0.2513 & 0.1586 & 0.1035 & 0.2078 & 0.7308 & 0.6042 & 0.3727 & 0.5096 & 0.7743 & 0.6278 & 0.3519 & 0.5223 \\
SimVG & BEiT-3-L & 0.3616 & \underline{0.2665} & \underline{0.2094} & \underline{0.3184} & 0.5596 & 0.3844 & 0.1862 & 0.3789 & 0.6537 & 0.4822 & 0.2977 & 0.4520 \\
LGR-NET & Swin-B & 0.3582 & 0.0883 & 0.0068 & 0.2480 & 0.5283 & 0.2308 & 0.0514 & 0.3181 & 0.5235 & 0.2419 & 0.0429 & 0.3186 \\
\midrule
\rowcolor[gray]{.95} \multicolumn{14}{l}{\textit{Baselines}} \\
OneRef & BEiT-3-B & 0.2684 & 0.1620 & 0.0976 & 0.2266 & 0.7575 & 0.6982 & 0.5524 & 0.5930 & 0.8333 & 0.7896 & 0.6448 & 0.6610 \\
OneRef & BEiT-3-L & 0.1323 & 0.0381 & 0.0073 & 0.1353 & 0.7594 & 0.7113 & 0.5947 & 0.6131 & 0.8366 & 0.8018 & 0.6942 & 0.6815 \\
\midrule

\rowcolor[gray]{.9} \textbf{VistaRef (Ours)} & BEiT-3-B & --- & --- & --- & --- & \textbf{0.8965} & \textbf{0.8190} & \textbf{0.6445} & \textbf{0.7077} & \textbf{0.9652} & \textbf{0.8875} & \textbf{0.6804} & \textbf{0.7518} \\
\rowcolor[gray]{.9} \textbf{VistaRef (Ours)} & BEiT-3-L & --- & --- & --- & --- & \textcolor{red}{\textbf{0.8836}} & \textcolor{red}{\textbf{0.8221}} & \textcolor{red}{\textbf{0.6842}} & \textcolor{red}{\textbf{0.7113}} & \textcolor{red}{\textbf{0.9701}} & \textcolor{red}{\textbf{0.9393}} & \textcolor{red}{\textbf{0.8261}} & \textcolor{red}{\textbf{0.8201}} \\
\bottomrule
\end{tabular}
}
\end{table*}
%%%%%%%%%%%%%%%%%%%%%%%%%%%%%%%%%%%%%%%%%%%%%%%%%%%%%%%%%%%%%%%%%%%%%%%%

%%%%%%%%%%%%%%%%%%%%%%%%%%%%%%%%%%%%%%%%%%%%%%%%%%%%%%%%%%%%%%%%%%%%%%%%%%%%%
\subsection{Local Hand Entity Modeling}
\label{sec:Local Hand Entity Modeling}
To capture fine-grained visual cues of the pointing source, the \textbf{LHEM} module extracts pure gesture representations from features by constructing the Local Hand Region $\Omega_{hand}$.\\
\textbf{Adaptive Hand Anchor.} 
$\Omega_{hand}$ is an axis-aligned rectangular region centered at $\mathbf{c}$. The localization of $\mathbf{c}$ follows a geometric prior:
\begin{equation}
\mathbf{c} = \mathbf{p}_r + \lambda_{center} \cdot \Delta
\end{equation}
where $\lambda_{center} = 0.3$. In the normalized coordinate system $[0, 1]$, the width of a single image patch is defined as $1/GridSize$. To ensure sufficient patch coverage even at extreme scales, we introduce a Minimum Spatial Prior $\epsilon_{min} = 2.0/GridSize$. The dimensions of the region are dynamically scaled based on the ray length $l$:
\begin{equation}
h_{long} = \max(\lambda_{long} \cdot l, \epsilon_{min}), \quad h_{short} = \max(\lambda_{short} \cdot l, \epsilon_{min})
\end{equation}
In our experiments, we set $\lambda_{long}=2.2$ and $\lambda_{short}=1.0$.\\
\textbf{Fine-grained Visual Sampling.} 
Let the raw visual feature sequence output by the backbone be $\mathbf{F}_v \in \mathbb{R}^{N \times d}$. For each patch $i$ located at $\mathbf{g}_i$, we construct a binary mask $M_i \in \{0, 1\}$ to anchor the hand region according to whether it falls within $\Omega_{hand}$:
\begin{equation}
\mathbf{M}_i = {1}(\mathbf{g}_i \in \Omega_{hand})
\end{equation}
If $\mathbf{M}_i = 0$ (i.e., no valid patches exist within the region), a Fallback Mechanism is employed to select the patch closest to the center $\mathbf{c}$ as the sampling result.\\

\textbf{Localized Query Synthesis.} 
To extract a pure gesture representation from complex visual scenes, we perform Masked Average Pooling on the features within the adaptive hand anchor. This process calculates the local hand feature vector $\mathbf{f}_{hand} \in \mathbb{R}^d$ as follows:
\begin{equation}
\mathbf{f}_{hand} = \frac{\sum_{i=1}^N \mathbf{M}_i \cdot \mathbf{F}_{v,i}}{\sum_{i=1}^N \mathbf{M}_i + \delta}
\end{equation}
where $\mathbf{M}_i \cdot \mathbf{F}_{v,i}$ utilizes the binary mask to accurately determine if the $i$-th patch resides within the anchor, thereby isolating pure hand patches. We introduce a minimal constant $\delta = 10^{-6}$ in the denominator to ensure numerical stability and further prevent computational collapse due to a zero denominator in extreme cases. Through this operation, the model compresses redundant regional information into a highly condensed semantic vector, providing a core gesture prior for subsequent query enhancement.

%%%%%%%%%%%%%%%%%%%%%%%%%%%%%%%%%%%%%%%%%%%%%%%%%%%%%%%%%%%%%%%%%%%%%%%% 图4
\begin{figure*}[!t]
    \centering
    \includegraphics[width=\linewidth]{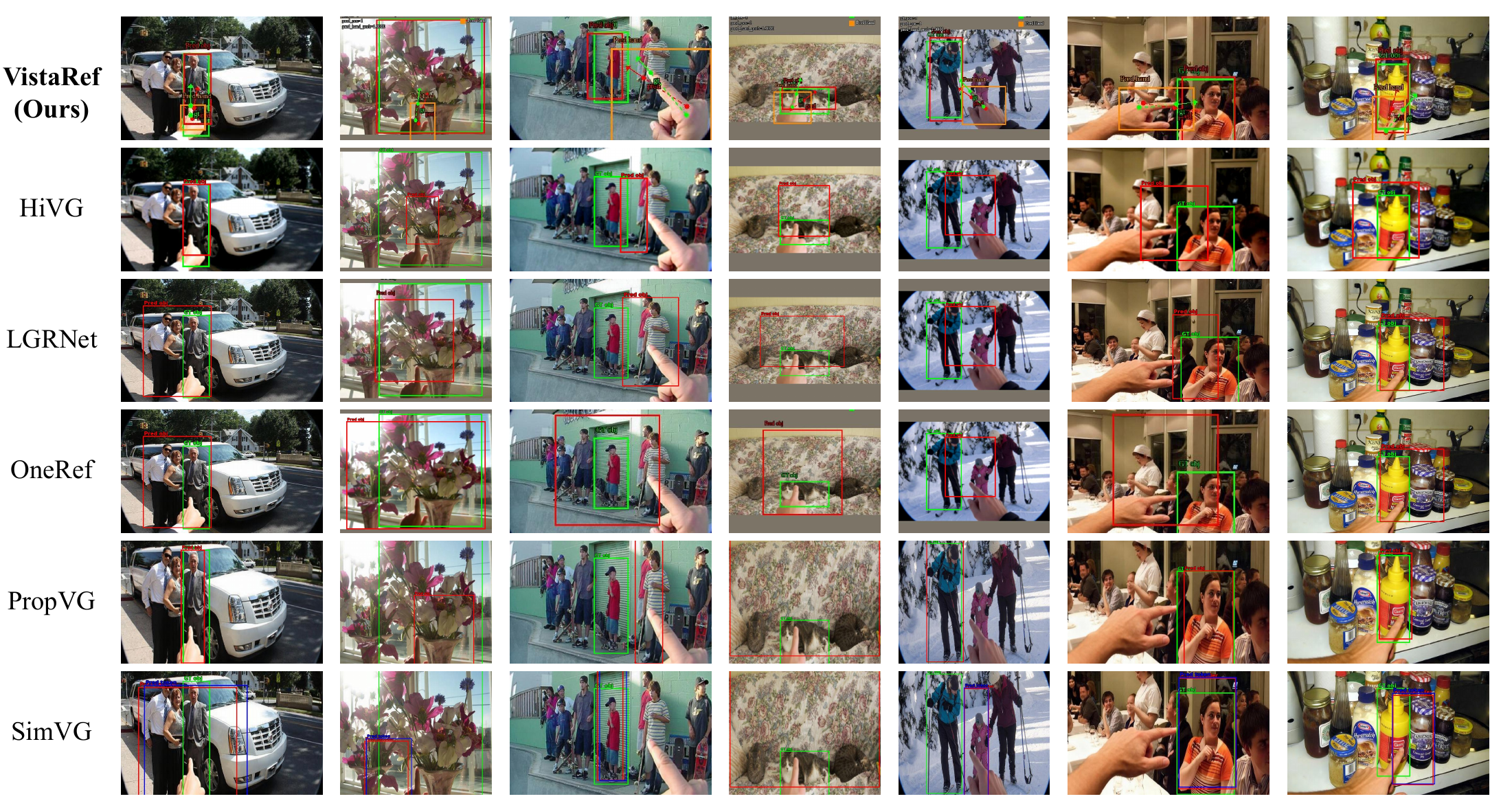}
    \caption{Visualization of grounding results. We visualize the results of VistaRef (top row) against several state-of-the-art baselines. VistaRef consistently achieves the most precise localization (Red boxes) across diverse indoor and outdoor scenarios, closely aligning with the ground truth (green boxes). Notably, in cluttered environments where baselines (e.g., SimVG, PropVG) suffer from severe orientation drift or predictive ambiguity, VistaRef effectively leverages explicit geometric rays to bridge the spatial perception gap, demonstrating superior orientation awareness.}
    \label{fig:Result-fig}
\end{figure*}
\vspace{-10pt}
%%%%%%%%%%%%%%%%%%%%%%%%%%%%%%%%%%%%%%%%%%%%%%%%%%%%%%%%%%%%%%%%%%%%%%%%
%%%%%%%%%%%%%%%%%%%%%%%   3.4、 射线引导的交互建模与目标回归
\vspace{1.0em} % 增加约一行的垂直间距

\subsection{Ray-guided Interaction Modeling and Box Regression}
\label{sec:Ray-guided Interaction Modeling and Box Regression}
To establish the semantic association between the pointing gesture and the target entity, we design a ray-guided interaction modeling module. First, a query fusion layer $MLP_{fuse}$ aggregates the finger-pointing ray prior $\mathbf{e}_{ray}$ and the local hand representation $\mathbf{f}_{hand}$ to synthesize an explicit query vector $\mathbf{Q}_{rh}$:
\begin{equation}
\mathbf{Q}_{rh} = MLP_{fuse}([\mathbf{f}_{hand}; \mathbf{e}_{ray}])
\end{equation}
This query vector $\mathbf{Q}_{rh}$ encapsulates both the geometric orientation and the visual appearance attributes of the hand. Subsequently, we employ a cross-attention mechanism to model the interaction between the gesture and the target.Taking $\mathbf{Q}_{rh}$ as the query and the global visual features $\mathbf{F}_v$ as both the key vector $\mathbf{K}_{F_v}$ and value vector $\mathbf{V}_{F_v}$, we compute the interaction feature $\mathbf{t}$:
\begin{equation}
\mathbf{t} = \text{CrossAttn}(\mathbf{Q}_{rh}, \mathbf{V}_{Fv}, \mathbf{K}_{Fv})
\end{equation}
The resulting target-aware feature $\mathbf{t}$ inherently captures the spatial association between the finger-pointing and the target object. We construct the decision feature $\mathbf{F}_{final}$ for bounding box regression. To leverage multi-scale information, we perform a non-linear fusion of the global semantic features $\mathbf{h}_{vl}$, the geometric prior $\mathbf{e}_{ray}$, and the interaction features $\mathbf{t}$ through concatenation:
\begin{equation}
\mathbf{F}_{final} = [\mathbf{h}_{vl}; \gamma \cdot \mathbf{e}_{ray}; \mathbf{t}]
\end{equation}
where $\mathbf{h}_{vl}$ provides the vision-language global context, $\gamma \cdot \mathbf{e}_{ray}$ offers gated geometric guidance, and $\mathbf{t}$ focuses on the filtered target candidate regions. Finally, $\mathbf{F}_{final}$ is fed into the bounding box regression head $MLP_{bbox}$ to predict the precise target coordinates $\mathbf{b} = [x_{center}, y_{center}, w, h] \in [0, 1]^4$:
\begin{equation}
\mathbf{b} = \sigma(MLP_{bbox}(\mathbf{F}_{final}))
\end{equation}
Here, $\gamma$ acts as a gated scaling factor (empirically set to 6.0) to modulate the geometric prior, while $x_{center}$ and $y_{center}$ denote the normalized center coordinates, and $w$ and $h$ represent the normalized width and height relative to the image dimensions. This design cascades reasoning from geometric priors to local interactions and global semantic fusion, enhancing localization robustness.

%%%%%%%%%%%%%%%%%%%%%%%  3.5 训练目标与条件监督
\subsection{Training Objectives and Conditional Supervision}
\label{sec:Training Objectives and Conditional Supervision}
During the training phase, we employ a multi-task joint optimization scheme to synergistically supervise the proposed architecture. The final total loss function $\mathcal{L}_{total}$ is defined as a weighted sum of the base detection loss and our proposed OCAL:

\begin{equation}
\mathcal{L}_{total} = \lambda_{base} \mathcal{L}_{base} + \lambda_{ocal} \mathcal{L}_{ocal}
\end{equation}

The base loss $\mathcal{L}_{base}$ follows the standard object detection framework to supervise target localization, which is formulated as:

\begin{equation}
\mathcal{L}_{base} = \mathcal{L}_{bbox} + \mathcal{L}_{giou}
\end{equation}

where $\mathcal{L}_{bbox}$ and $\mathcal{L}_{giou}$ denote the bounding box regression loss and the Generalized Intersection over Union loss (GIoU) \cite{rezatofighi2019generalized}, respectively. To explicitly model the "point-to-line" physical process, the OCAL $\mathcal{L}_{ocal}$ is introduced to supervise hand entity perception and geometric consistency. For positive samples containing valid pointing gestures, $\mathcal{L}_{ocal}$ is defined as:

\begin{equation}
\mathcal{L}_{ocal} = \lambda_{hand} \mathcal{L}_{hand} + \lambda_{kp} \mathcal{L}_{kp} + \lambda_{ray} \mathcal{L}_{ray}
\end{equation}

Specifically, $\mathcal{L}_{hand}$ is responsible for monitoring the presence of the hand entity, $\mathcal{L}_{kp}$ enables fine-grained geometric perception of the hand keypoints, and $\mathcal{L}_{ray}$ serves as the core orientation ray constraint. To ensure the purity of the ray-aware interaction logic, we implement an asymmetric supervision mechanism based on the results from the hand detection head $\mathbf{Head}_{pred}$. Specifically, if $\mathbf{Head}_{pred}$ determines that no pointing gesture is present in the image, we only retain the hand classification loss $\mathcal{L}_{hand}$ to supervise the presence of the entity, while the coefficients for the geometric components, $\lambda_{kp}$ and $\lambda_{ray}$, are set to zero.

%%%%%%%%%%%%%%%%%%%%%%%%%%%%%%%%%%%%  4、Main Results

\section{Experiments}

%%%%%%%%%%%%%%%%%%%%%%%%%%%%%%%表格2 module 消融%%%%%%%%%%%%%%%%%%%%%%%%%
% \begin{table*}[t]
% \centering
% \caption{\textbf{Ablation study of VistaRef components.} We investigate the impact of Local Hand Entity Modeling (LHEM), Geometric Ray Modeling (GRM), and Cross-Attention on the final performance. All experiments are conducted using the BEiT-3-Base backbone.}
% \label{tab:ablation}
% \begin{tabular}{ccccc|cc|cc}
% \toprule
% \textbf{Exp.} & \textbf{Baseline} & \textbf{LHEM} & \textbf{GRM} & \textbf{Cross-Attention} & \textbf{P@0.3} $\uparrow$ & \textbf{P@0.5} $\uparrow$ & \textbf{P@0.7} $\uparrow$ & \textbf{mIoU} $\uparrow$ \\
% \midrule
% (a) & \checkmark & & & & 0.7578 & 0.6973 & 0.5557 & 0.5939 \\
% (b) & \checkmark & \checkmark & & & 0.8838 & 0.7891 & 0.5892 & 0.6839 \\
% (c) & \checkmark & & \checkmark & & 0.7568 & 0.6768 & 0.5241 & 0.5754 \\
% (d) & \checkmark & \checkmark & \checkmark & & \underline{0.8887} & \underline{0.7816} & \underline{0.5918} & \underline{0.6824} \\
% (e) & \checkmark & \checkmark & \checkmark & \checkmark & \textbf{0.8920} & \textbf{0.7975} & \textbf{0.6097} & \textbf{0.6919} \\
% \bottomrule
% \end{tabular}
% \end{table*}

\begin{table*}[t]
\centering
\caption{\textbf{Ablation study of VistaRef components.} We investigate the impact of Geometric Ray Modeling (GRM), Local Hand Entity Modeling (LHEM), and Cross-Attention on the final performance. All experiments are conducted using the BEiT-3-Base backbone.}
\label{tab:ablation}
\begin{tabular}{ccccc|cc|cc}
\toprule
\textbf{Exp.} & \textbf{Baseline} & \textbf{GRM} & \textbf{LHEM} & \textbf{Cross-Attention} & \textbf{P@0.3} $\uparrow$ & \textbf{P@0.5} $\uparrow$ & \textbf{P@0.7} $\uparrow$ & \textbf{mIoU} $\uparrow$ \\
\midrule
(a) & \checkmark & & & & 0.7578 & 0.6973 & 0.5557 & 0.5939 \\
(b) & \checkmark & \checkmark & & & 0.7568 & 0.6768 & 0.5241 & 0.5754 \\
(c) & \checkmark & & \checkmark & & 0.8838 & 0.7891 & 0.5892 & 0.6839 \\
(d) & \checkmark & \checkmark & \checkmark & & \underline{0.8887} & \underline{0.7816} & \underline{0.5918} & \underline{0.6824} \\
(e) & \checkmark & \checkmark & \checkmark & \checkmark & \textbf{0.8920} & \textbf{0.7975} & \textbf{0.6097} & \textbf{0.6919} \\
\bottomrule
\end{tabular}
\end{table*}

%%%%%%%%%%%%%%%%%%%%%%%%%%%%%%%%%%%%%%%%%%%%%%%%%%%%%%%%%%%%%%%%%%%%%%%%
%%%%%%%%%%%%%%%%%%%%%%%%%%%%%%%%%%%loss消融%%%%%%%%%%%%%%%%%%%%%%%%%%%%%%
% \begin{table}[t]
% \centering
% \caption{\textbf{Ablation study on the components of OCAL.} We investigate the contribution of each supervision term within the Orientation-Consistent Alignment Loss (OCAL). $\mathcal{L}_{hand}$, $\mathcal{L}_{kp}$, and $\mathcal{L}_{ray}$ denote the hand presence, keypoint localization, and orientation ray losses, respectively.}
% \label{tab:loss_ablation}
% \begin{tabular}{c|ccc|cccc}
% \toprule
% \textbf{Exp.} & $\mathcal{L}_{hand}$ & $\mathcal{L}_{kp}$ & $\mathcal{L}_{ray}$ & \textbf{P@0.3} $\uparrow$ & \textbf{P@0.5} $\uparrow$ & \textbf{P@0.7} $\uparrow$ & \textbf{mIoU} $\uparrow$ \\
% \midrule
% Baseline & & & & 0.5924 & 0.5479 & 0.4339 & 0.4611 \\
% \midrule
% (a) & \checkmark & & & 0.8441 & 0.7077 & 0.4352 & 0.6165 \\
% (b) & & \checkmark & & 0.7490 & 0.6579 & 0.4548 & 0.5483 \\
% (c) & & & \checkmark & 0.7513 & 0.6755 & 0.5055 & 0.5673 \\
% (d) & \checkmark & \checkmark & & \underline{0.8776} & \underline{0.7718} & \underline{0.5485} & \underline{0.6663} \\
% \midrule
% \textbf{Full} & \checkmark & \checkmark & \checkmark & \textbf{0.8920} & \textbf{0.7975} & \textbf{0.6097} & \textbf{0.6919} \\
% \bottomrule
% \end{tabular}
% \end{table}
%%%%%%%%%%%%%%%%%%%%%%%%%%%%%%   表3 loss消融
\begin{table}[t]
\centering
\small % 基础字号调小
\caption{\textbf{Ablation study on OCAL components.} $\mathcal{L}_{hand}$, $\mathcal{L}_{kp}$, and $\mathcal{L}_{ray}$ represent hand presence, keypoint, and orientation ray losses, respectively. $\uparrow$ indicates higher values are better.}
\label{tab:loss_ablation}
\setlength{\tabcolsep}{3.5pt} % 关键：手动缩减列间距（默认通常是 6pt）
\resizebox{\columnwidth}{!}{% % 关键：强制适配单栏宽度
\begin{tabular}{c|ccc|cccc}
\toprule
\textbf{Exp.} & $\mathcal{L}_{hand}$ & $\mathcal{L}_{kp}$ & $\mathcal{L}_{ray}$ & \textbf{P@0.3} & \textbf{P@0.5} & \textbf{P@0.7} & \textbf{mIoU} \\
\midrule
Baseline & & & & 0.5924 & 0.5479 & 0.4339 & 0.4611 \\
\midrule
(a) & \checkmark & & & 0.8441 & 0.7077 & 0.4352 & 0.6165 \\
(b) & & \checkmark & & 0.7490 & 0.6579 & 0.4548 & 0.5483 \\
(c) & & & \checkmark & 0.7513 & 0.6755 & 0.5055 & 0.5673 \\
(d) & \checkmark & \checkmark & & \underline{0.8776} & \underline{0.7718} & \underline{0.5485} & \underline{0.6663} \\
\midrule
\textbf{Full} & \checkmark & \checkmark & \checkmark & \textbf{0.8920} & \textbf{0.7975} & \textbf{0.6097} & \textbf{0.6919} \\
\bottomrule
\end{tabular}%
}
\end{table}
\begin{table}[t]
\centering
\footnotesize % 1. 使用较小字号作为基准
\caption{\textbf{Sensitivity analysis of loss weights.} We evaluate the impact of weight variations for $\mathcal{L}_{kp}$, $\mathcal{L}_{ray}$, and $\mathcal{L}_{hand}$. $\uparrow$ denotes higher values are better.}
\label{tab:parameter_sensitivity}
\setlength{\tabcolsep}{3.5pt} % 2. 核心：大幅缩减列与列之间的间距（默认为6pt）
\resizebox{\columnwidth}{!}{% % 3. 强制适配当前栏宽
\begin{tabular}{lccccc} % 第一列改用左对齐 'l'，视觉上更整齐
\toprule
\textbf{Comp.} & \textbf{Weight} & \textbf{P@0.3} & \textbf{P@0.5} & \textbf{P@0.7} & \textbf{mIoU} \\ 
\midrule
\multirow{3}{*}{$\mathcal{L}_{kp}$}   & 0.2 & 0.8880 & \textbf{0.8043} & \textbf{0.6259} & \textbf{0.6988} \\
                                      & 0.4 & \textbf{0.8920} & 0.7975 & 0.6097 & 0.6919 \\
                                      & 0.6 & 0.8857 & 0.7848 & 0.5647 & 0.6726 \\ 
\midrule
\multirow{3}{*}{$\mathcal{L}_{ray}$}  & 0.2 & 0.8896 & \textbf{0.8014} & 0.6074 & \textbf{0.6919} \\
                                      & 0.4 & \textbf{0.8920} & 0.7975 & \textbf{0.6097} & \textbf{0.6919} \\
                                      & 0.6 & 0.8912 & 0.8046 & 0.6308 & 0.7004 \\ 
\midrule
\multirow{3}{*}{$\mathcal{L}_{hand}$} & 0.1 & 0.8916 & \textbf{0.8102} & \textbf{0.6471} & \textbf{0.7050} \\
                                      & 0.2 & \textbf{0.8920} & 0.7975 & 0.6097 & 0.6919 \\
                                      & 0.3 & 0.8750 & 0.7487 & 0.4847 & 0.6445 \\ 
\bottomrule
\end{tabular}
}
\end{table}
\vspace{-2pt}
%%%%%%%%%%%%%%%%%%%%%%%%%%%%%%%%%%%%%%%%%%%%%%%%%%%%%%%%%%%%%%%%%%%%%%%
% \begin{table}[t]
% \centering
% \caption{\textbf{Sensitivity analysis of loss weights for different components.} We evaluate the impact of different weight values for $\mathcal{L}_{kp}$, $\mathcal{L}_{ray}$, and $\mathcal{L}_{hand}$. The missing values are denoted by 0.xxxx.}
% \label{tab:parameter_sensitivity}
% \begin{tabular}{lccccc}
% \toprule
% \textbf{Component} & \textbf{Value} & \textbf{P@0.3} $\uparrow$ & \textbf{P@0.5} $\uparrow$ & \textbf{P@0.7} $\uparrow$ & \textbf{mIoU} $\uparrow$ \\ 
% \midrule
% \multirow{3}{*}{\mathcal{L}_{kp}$}   & 0.2 & 0.xxxx & 0.xxxx & 0.xxxx & 0.xxxx \\
%                             & 0.4 & 0.8920 & 0.7975 & 0.6097 & 0.6919 \\
%                             & 0.6 & 0.xxxx & 0.xxxx & 0.xxxx & 0.xxxx \\ 
% \midrule
% \multirow{3}{*}{\mathcal{L}_{ray}$}  & 0.2 & 0.8896 & 0.8014 & 0.6074 & 0.6919 \\
%                             & 0.4 & 0.8920 & 0.7975 & 0.6097 & 0.6919 \\
%                             & 0.6 & 0.xxxx & 0.xxxx & 0.xxxx & 0.xxxx \\ 
% \midrule
% \multirow{3}{*}{\mathcal{L}_{hand}$} & 0.1 & 0.8916 & 0.8102 & 0.6471 & 0.7050 \\
%                             & 0.2 & 0.8920 & 0.7975 & 0.6097 & 0.6919 \\
%                             & 0.3 & 0.xxxx & 0.xxxx & 0.xxxx & 0.xxxx \\ 
% \bottomrule
% \end{tabular}
% \end{table}

\subsection{Experimental Settings}
%%%%%%%%%%%%%TO DO BY CAIZHIZHEN%%%%%%%%%%%%%%%%%%%%
We evaluate our framework on the EgoPoint-Ground dataset \cite{li2026languagegroundingreferringexpressions}, a benchmark specifically designed for egocentric pointing-based target localization. Diverging from conventional visual grounding, this task emphasizes gesture-based disambiguation, requiring the model to localize the object indicated by a pointing finger given a fixed textual prompt. Each image provides annotations for hand instances, target objects, and finger keypoints (root and fingertip). During preprocessing, the hand instance is isolated from other object annotations, which serve as candidate targets. Negative samples are assigned a null bounding box and a binary \texttt{is\_positive} label. The textual query is standardized as: ``Please frame the bbox where a finger points to the item.'' Performance is quantified using mean Intersection over Union (mIoU) and Precision@$k$ ($k \in \{0.3, 0.5, 0.7\}$). Specifically, Precision@0.5 serves as the primary metric, while Precision@0.3 and Precision@0.7 characterize coarse and high-precision localization, respectively. All results are reported on the test set. VistaRef optimized via a 10-epoch warm-up ($LR=2.5 \times 10^{-4}$) followed by a 20-epoch fine-tuning stage ($LR=3 \times 10^{-5}$). The training objective is balanced by weights $\lambda_{base}=0.7$ and $\lambda_{ocal}=0.3$, further stabilized using an EMA-based loss normalization strategy with $0.99$ momentum. Refer to the Appendix for more details.

\vspace{-10pt}
%%%%%%%%%%%%%%%%%%%%%%%%%%%%%%%%%%%%%%%%%%%%%%%%%%%%%%%%%%%%%%%%%%%%%%%%
\subsection{Main Results}
As illustrated in  Table \ref{tab:main_results}, we conducted comparative experiments on the pointing-to-object detection task under three distinct evaluation protocols: Zero-shot transfer, Training on Hybrid, and Training-Real Data Test. Qualitatively, VistaRef demonstrates superior performance in resolving complex spatial ambiguities compared to state-of-the-art baselines. As illustrated in Fig. \ref{fig:Result-fig}, while traditional attention-based models (e.g., SimVG, PropVG) often suffer from significant localization drift—particularly in cluttered environments or when dealing with distant targets—VistaRef maintains deterministic ray alignment by explicitly modeling the geometric correlation from hand to target. By leveraging its Geometric Ray Modeling (GRM) and Local Hand Entity Modeling (LHEM) modules, our framework effectively bridges the "spatial perception gap" and overcomes "orientation insensitivity." This allows for high-precision grounding even when faced with subtle finger pose variations or dense distractors, ensuring robust and stable target localization across diverse and challenging interaction scenarios.

\textbf{Zero-shot Analysis.} Without the pointing-to-object detection task training, conventional VG models exhibit substandard performance. The best-performing model, SimVG \cite{dai2024simvgsimpleframeworkvisual} (BEiT-3-L), achieves only 26.65\% (P@0.5), 20.94\% (P@0.7), and 31.84\% (mIoU). These models are primarily trained on purely linguistic referential expressions that exclude deictic signals, such as "\textit{A baseball player in the middle of a throwing motion}''. When directly transferred to tasks necessitating gesture-based guidance, such as "\textit{A finger pointing at a baseball player in the middle of a throwing motion}'' (as shown in Fig.~\ref{fig:vista_ref_overview}), these models effectively recognize object semantics but fail to comprehend the directional cues linking the finger to the target object. This inability to interpret deictic signals hinders the model's capacity to distinguish between multiple targets sharing identical or similar features within a scene (as shown in Fig.~\ref{fig:Result-fig}).\\
\textbf{Hybrid Data Performance.} Training on the hybrid dataset incorporating the RefCOCO series \cite{yu2016refcoco, mao2016refcoco+, nagaraja2016refcocog} and generative samples—yields consistent gains across all models. Among REC baselines, PropVG \cite{dai2025propvgendtoendproposaldrivenvisual} (BEiT-3-L) performs best, yet VistaRef (Ours) establishes a new state-of-the-art across all metrics. Notably, even our base variant (BEiT-3-B) significantly outperforms PropVG (BEiT-3-L) by 9.99\% in P@0.5 and 6.36\% in mIoU, with the large variant further pushing P@0.5 to 82.21\%. This superior efficacy stems from our LHEM and GRM modules (Fig.~\ref{fig:ray_hand_modeling}), which explicitly encode gestural geometry, and the Ray-Aware Target Grounding module that enables precise localization along physical pointing rays. In contrast, traditional VG models \cite{Xiao_2024, dai2024simvgsimpleframeworkvisual, xiao2024onerefunifiedonetowerexpression} struggle to resolve the physical implications of gestures due to their reliance on implicit semantic matching rather than explicit geometric parsing.\\
\textbf{Real-world Data Evaluation.} In the Train-Real Data Test task, characterized by higher precision in real-world finger pointing, the mIoU of VistaRef (BEiT-3-B) increases from its Hybrid environment baseline of 70.77\% to 75.18\%, yielding a gain of approximately 4.41\%. Similarly, the mIoU for VistaRef (BEiT-3-L) escalates from 71.13\% to 82.01\%, marking an increase of approximately 10.88\%. On the pivotal mIoU metric, both versions of VistaRef maintain a 4.7\% lead over the strongest conventional baseline, PropVG \cite{dai2025propvgendtoendproposaldrivenvisual} (BEiT-3-L) (77.45\%). These results demonstrate that VistaRef maintains superior localization accuracy through its robust geometric reasoning, even in complex real-world scenes populated by multiple objects with similar visual characteristics.
% \vspace{-2pt}
%%%%%%%%%%%%%%%%%%%%%%%%%%%%%%%%%%%%%%%%%%%%%%%%%% 表格5：模型效率对比
\begin{table*}[!t]
\centering
\caption{\textbf{Comparison of model efficiency and parameters.} All experiments are conducted on the BEiT-3-Base backbone with a total of 3,072 samples.}
\label{tab:efficiency_comparison}
\begin{tabular}{lcccc}
\toprule
\textbf{Method} & \textbf{Total Params (M)} & \textbf{Trainable Params (M)} & \textbf{Total Time (s)} & \textbf{Throughput (fps)} \\ 
\midrule
PropVG\cite{dai2025propvgendtoendproposaldrivenvisual} & 246.82 & 246.82 & 431.70 & 7.12 \\
SimVG\cite{dai2024simvgsimpleframeworkvisual}  & 230.16 & 230.16 & 130.22 & 23.59 \\
HiVG\cite{Xiao_2024}   & 213.24 & 63.62  & 119.01 & 25.81 \\
LGRNet\cite{lu2024lgr} & 335.07 & 225.59 & 112.77 & 27.24 \\
OneRef\cite{xiao2024onerefunifiedonetowerexpression} & 224.82 & 224.82 & 92.31  & 33.28 \\
\midrule
\textbf{VistaRef (Ours)} & 226.01 & 226.01 & 134.01 & 22.92 \\ 
\bottomrule
\end{tabular}
\end{table*}
%%%%%%%%%%%%%%%%%%%%%%%%%%%%%%%%%%%%%%%%%%%%%%%%%%%%%
%%%%%%%%%% 4.3 消融实验%%%%%%%%%%%%%%%%%%%%%%%%%%%%%%%%%%%%%%%%%%%%%%%%%
% \vspace{-16pt}
\subsection{Ablation Study}

To evaluate the effectiveness of the core modules in VistaRef and the effect of each individual loss in OCAL, we conducted a series of ablation studies using OneRef \cite{xiao2024onerefunifiedonetowerexpression} as the baseline. The experimental results are show in Table \ref{tab:ablation} and Table \ref{tab:loss_ablation}.

% Table \ref{tab:ablation} illustrates the individual effectiveness of each core module within VistaRef, specifically GRM, LHEM, and the Cross-Attention mechanism.
% Comparing experiment (c) (mIoU of $68.39\%$), experiment (d) ($68.24\%$), and experiment (a) ($59.39\%$) in Table \ref{tab:ablation} indicates that the GRM and LHEM modules are essential for strengthening the model's capability to understand the spatial implications of pointing gestures.
% Furthermore, the comparison between experiment (d) ($68.24\%$) and (e) ($69.19\%$) validates the critical role of the cross-attention mechanism in enhancing grounding accuracy. In experiment (e), the $\mathbf{Q}_{rh}$ is generated by fusing the geometric orientation $\mathbf{e}_{ray}$ (produced by GRM) with the local hand representation $\mathbf{f}_{hand}$ (produced by LHEM). This query is then used to retrieve relevant entities within the global image features $\mathbf{F}_v$, as illustrated in Figure \ref{fig:vista_ref_overview}. The resulting interaction feature $\mathbf{t}$ deeply encodes the spatial associations, which are ultimately combined with the global semantic feature $\mathbf{h}_{vl}$ and the geometric prior $\mathbf{e}_{ray}$ to construct the decision feature $\mathbf{F}_{final}$. Ultimately, $\mathbf{F}_{final}$ effectively guides the Ray-Guided Target Grounding process, eliminating localization ambiguities in complex scenarios and significantly enhancing the model's physical spatial reasoning capabilities.
Table \ref{tab:ablation} quantifies the performance contribution of each core component, including GRM, LHEM, and the Cross-Attention mechanism. The baseline model (exp. a) achieves an mIoU of only $59.39\%$, underscoring the inherent difficulty of resolving pointing intent without explicit spatial constraints. The introduction of either GRM (exp. c, $68.39\%$) or LHEM (exp. d, $68.24\%$) yields a substantial performance leap of approximately 9\% in mIoU, demonstrating that explicit geometric modeling of the pointing ray and gestural cues is fundamental to spatial disambiguation. The progression to experiment (e) ($69.19\%$ mIoU) further validates the synergy achieved via the Cross-Attention mechanism. By effectively fusing hand-pose embeddings with geometric orientation, the model generates a high-fidelity joint query that retrieves spatially relevant entities from global features. This integration ensures that the final decision process is guided by physically consistent spatial reasoning, thereby eliminating localization drift in complex, cluttered scenarios.

%%%%%%%%%%%%%%%%%%%%%%%%%%%%%%%  表格3 损失——消融

Table \ref{tab:loss_ablation} evaluates the individual contribution of each supervisory loss term in OCAL: hand classification loss ($\mathcal{L}_{hand}$), keypoint regression loss ($\mathcal{L}_{kp}$), and orientation ray loss ($\mathcal{L}_{ray}$). Compared to the baseline (46.11\% mIoU), adding $\mathcal{L}_{hand}$ in Exp. (a) yields a substantial surge to 61.65\%, confirming that hand presence detection is a fundamental prerequisite for accurate localization. Without $\mathcal{L}_{hand}$, experiment (b) and (c) show limited gains (54.83\% and 56.73\% mIoU) as the model becomes susceptible to interference from negative samples lacking pointing gestures. Finally, the full OCAL framework, integrating all losses, achieves optimal 69.19\% mIoU, demonstrating comprehensive supervision enables more robust structural gesture feature extraction.

%%%%%%%%%%%%%%%%%%%%%%%%%%%%%%%%%%%%  4.3 loss 超参数分析
\vspace{-15pt}
\subsection{Hyperparameter Sensitivity Analysis}
The sensitivity analysis in Table \ref{tab:parameter_sensitivity} validates the remarkable robustness of VistaRef against variations in loss weights. Experimental data indicate that as the weight of $\mathcal{L}_{kp}$ increases from 0.2 to 0.6, the P@0.3 metric exhibits only a marginal fluctuation between $0.8880$ and $0.8857$ (a negligible decline of less than 0.3\%). Furthermore, within the tested range for $\mathcal{L}_{ray}$, P@0.3 steadily improves from $0.8896$ to $0.8920$, while the mIoU remains remarkably stable at 0.6919, demonstrating the high predictive stability of the internal geometric alignment mechanism across varying supervision intensities. Although a slight performance convergence is observed when $\mathcal{L}_{hand}$ increases to $0.3$—likely due to an imbalance in supervision—the mIoU consistently stays above the 0.69 threshold within the core interval of $0.1$ to $0.2$. This characteristic of low parameter sensitivity underscores the framework's superior performance consistency and significantly alleviates the burden of hyperparameter tuning in practical deployments.

%%%%%%%%%%%%%%%%%%%%%%%%%%%%%%%%   4.5  推理效率分析
\vspace{-5pt}
\subsection{Model Complexity and Efficiency Analysis}
Table \ref{tab:efficiency_comparison} illustrates the comparison of model complexity and efficiency among various models on the BEiT-3-Base \cite{wang2022imageforeignlanguagebeit} backbone. In terms of model scale, the total parameter count of VistaRef (Ours) is 226.01M, which is comparable to the lightweight baseline OneRef \cite{xiao2024onerefunifiedonetowerexpression} (224.82M) and significantly lower than LGRNet (335.07M). This demonstrates that the geometric modeling modules introduced in this study are highly concise and do not impose a substantial storage burden on the system. Regarding inference efficiency, VistaRef achieves a throughput of 22.92 fps when processing 3,072 samples. Compared to PropVG \cite{dai2025propvgendtoendproposaldrivenvisual} (7.12 fps), its execution speed is improved by approximately 3.2$\times$, exhibiting a distinct advantage in real-time performance. Although its speed is slightly lower than that of the baseline OneRef \cite{xiao2024onerefunifiedonetowerexpression}, this marginal increase in computational overhead is entirely acceptable given the precision improvements of VistaRef in pointing tasks within complex multi-object scenarios. In summary, without a significant increase in parameter count, VistaRef balances high-precision localization with superior computational efficiency, achieving an optimal trade-off between model performance and Inference efficiency.

\section{Conclusion}
This paper presents VistaRef, a framework specifically designed to address the inherent deficiencies of Transformer-based models in spatial orientation awareness. We enhance system performance through three core architectural innovations: LHEM to capture subtle finger articulations, GRM to represent the spatial trajectory from the hand to the target, and the Orientation-Consistent Alignment Loss to ensure physical self-consistency. Experimental results demonstrate that VistaRef effectively mitigates localization drift and significantly improves pointing accuracy in cluttered or distant scenarios.

\noindent\textbf{Limitations:}
Despite its performance, VistaRef has several core limitations. First is the precision bottleneck under physical constraints: when targets are extremely distant, or the hand region suffers from low pixel resolution, minute pose estimation errors are magnified over the distance, leading to substantial ray deviation. Second is the relative lack of semantic reasoning; currently, the model prioritizes the geometric ray path over intrinsic object attributes. Consequently, in scenarios where the ray intersects multiple overlapping objects, the system may select incorrect targets due to insufficient class-specific understanding. Addressing these challenges remains a primary focus for our future research.
\newpage
\appendix

\section{Dataset and Evaluation Metrics}
%%%%%%%%%%%%%TO DO%%%%%%%%%%%%%%
\subsubsection{Dataset Introduction}

We conduct experiments on an egocentric pointing-based visual grounding dataset constructed from publicly available image sources. 
Unlike conventional grounding benchmarks that primarily rely on linguistic descriptions, this dataset emphasizes \emph{gesture-driven disambiguation}, 
where the referent is specified implicitly via a pointing action.

Each image captures a natural interaction scenario in which a human hand points to a target object. 
The dataset provides annotations including hand and object bounding boxes, as well as fingertip keypoints (e.g., root and fingertip), 
enabling fine-grained modeling of spatial and geometric relationships.

During preprocessing, the hand instance is separated from other annotated objects, which are treated as candidate targets. 
For samples without valid pointing targets, we introduce a binary indicator \texttt{is\_positive} and set the target bounding box to zero. 
Following standard practice, we use a fixed textual query:
\begin{equation}
\texttt{Please frame the bbox where a finger points to the item.}
\end{equation}
This setting removes linguistic variability and focuses the task on spatial reasoning induced by pointing gestures.

The dataset contains 15,455 images, split into 10,837/1,546/3,072 for training, validation, and testing, respectively. 
covering 345 object categories with diverse scene complexity.

\subsubsection{Evaluation Metrics}

We adopt standard localization metrics based on Intersection over Union (IoU).

\textbf{Intersection over Union (IoU).}
Given a predicted bounding box $B_p$ and the ground-truth bounding box $B_{gt}$, IoU is defined as:
\begin{equation}
\text{IoU}(B_p, B_{gt}) = \frac{|B_p \cap B_{gt}|}{|B_p \cup B_{gt}|}.
\end{equation}

\textbf{Precision@$\tau$.}
We report Precision@$\tau$, defined as the proportion of predictions whose IoU exceeds a threshold $\tau$:
\begin{equation}
\text{Precision@}\tau = \frac{1}{N} \sum_{i=1}^{N} \mathbb{I} \left( \text{IoU}_i \geq \tau \right),
\end{equation}
where $N$ is the total number of samples and $\mathbb{I}(\cdot)$ denotes the indicator function. 
We evaluate at $\tau \in \{0.3, 0.5, 0.7\}$.

\textbf{Mean IoU (mIoU).}
We further report the mean IoU over all samples:
\begin{equation}
\text{mIoU} = \frac{1}{N} \sum_{i=1}^{N} \text{IoU}_i.
\end{equation}
%%%%%%%%%%%%%%%%%%%%%%%%%%%%%%%%%%%%%%%%%%%%%%%%%%%%%%%%%%%%%
\section{Model Details}
%%%%%%%%%%%%%%%%%%%%%%%%%%%%%%%%%%%%%%%%%%%%%%%
\begin{figure*}
    \centering
    \includegraphics[width=\linewidth]{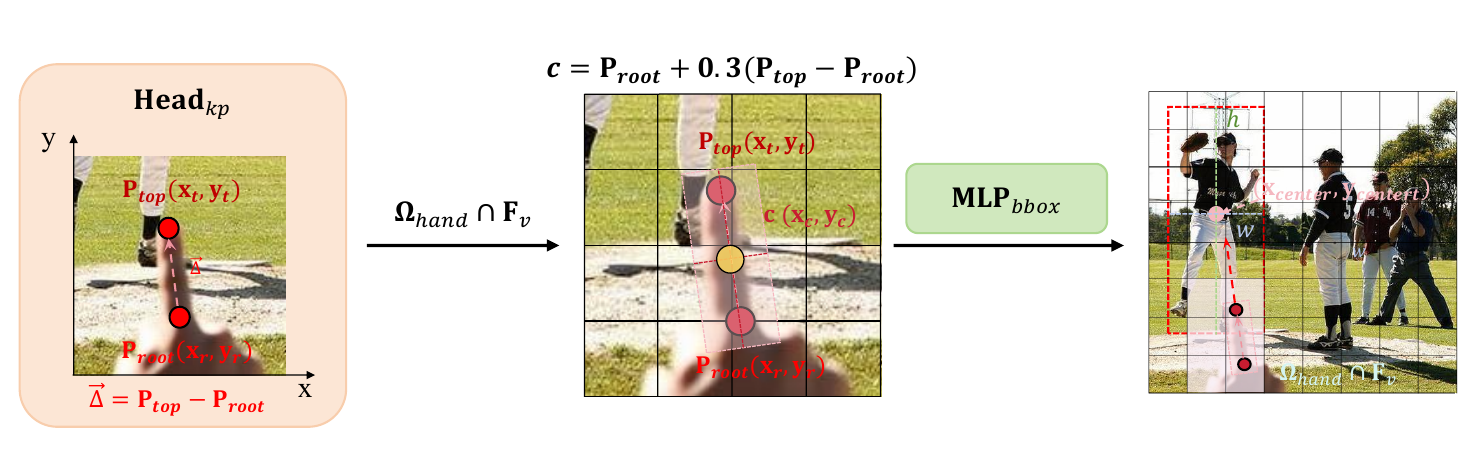}
    \caption{More detailed visualization of fingertip localization.}
    \label{fig:supple-hand}
\end{figure*}
%%%%%%%%%%%%%%%%%%%%%%%%%%%%%%%%%%%%%%%%%%%%%

\subsection{Architectural Configurations}
We provide an exhaustive description of the dimensional configurations for each component within \textbf{VistaRef}. All Multi-Layer Perceptron (MLP) modules follow a consistent three-layer architecture, consisting of two hidden layers with dimension $d$ (where $d=768$ for the Base model and $d=1024$ for Large) and ReLU activation functions, followed by a linear output projection.\\
\textbf{Keypoint Detection Head ($\mathbf{Head}_{kp}$):} Keypoint detection head receives the text-guided visual feature $\mathbf{h}_{vl} \in \mathbb{R}^d$ as input with dimension of $d$. It outputs a 4-dimensional vector which, through Sigmoid activation, represents the specific coordinates for the root $\mathbf{p}_r$ and tip $\mathbf{p}_t$ of the finger.\\
\textbf{Ray Encoder:} The ray encoder serves as a bridge between explicit geometric reasoning and high-dimensional feature alignment. It takes a 7-dimensional geometric primitive vector $\mathbf{g}_{ray} \in \mathbb{R}^7$ as input, which is explicitly constructed by concatenating the predicted hand root $\mathbf{p}_r$, fingertip $\mathbf{p}_t$, the normalized unit direction vector $\mathbf{u}$, and the scalar ray length $l$ (i.e., $\mathbf{g}_{ray} = [\mathbf{p}_r; \mathbf{p}_t; \mathbf{u}; l]$). Utilizing a three-layer MLP, the encoder projects these low-level geometric priors into the $d$-dimensional latent space, yielding the ray embedding $\mathbf{e}_{ray} \in \mathbb{R}^d$ that can be seamlessly integrated with visual and linguistic representations.\\
\textbf{Local Hand Query Fusion:} This layer utilizes a three-layer MLP to fuse the $d$-dimensional local hand feature $\mathbf{f}_{hand}$ with the $d$-dimensional $\mathbf{e}_{ray}$ through concatenation, resulting in an input dimension of $2d$. This $2d$ vector is then processed by a three-layer MLP, where the initial linear layer performs a dimensionality reduction from $2d$ to $d$, followed by a ReLU activation and subsequent hidden layers that maintain the $d$-dimensional space. The output is the $d$-dimensional explicit query vector $\mathbf{Q}_{rh}$.\\
\textbf{Cross-Attention Mechanism:} We employ a multi-head cross-attention mechanism with 8 heads. $\mathbf{Q}_{rh}$ serves as the \textit{Query}, while the global visual patch features $\mathbf{F}_v$ serve as the \textit{Key} ($\mathbf{K}_{\mathbf{F}_v}$) and \textit{Value} ($\mathbf{V}_{\mathbf{F}_v}$), resulting in the interaction feature $\mathbf{t}$.

%%%%%%%%%%%%% B2
\subsection{Training and Optimization}
\textbf{Warm-up Phase:} To prevent large gradient fluctuations caused by random weight initialization during the early stages of training, we implement a 10-epoch \textit{warm-up} period. During this phase, the learning rate increases linearly from a near-zero value to the target base learning rate, ensuring the model enters a stable optimization trajectory.\\
\textbf{Optimization Schedule:} Following the warm-up phase, the model undergoes 20 epochs of fine-tuning. The learning rate is adjusted according to a decay strategy, complemented by Exponential Moving Average (EMA) with a momentum of 0.99 to smooth the learned weights.\\
\textbf{Loss Balancing:} The total loss is composed of $L_{base}$ with a weight of 0.7 and $L_{ocal}$ with a weight of 0.3. Within the Orientation-Consistent Alignment Loss (OCAL) framework, the internal weights are distributed as $\lambda_{hand}=0.2, \lambda_{kp}=0.4, \text{ and } \lambda_{ray}=0.4$.

%%%%%%%%%%%%%%%%%% B 3  
\subsection{Data and Prompting}
\textbf{Prompt Engineering:} All experiments utilize a unified and standardized textual instruction: \textit{``Please frame the bbox where a finger points to the item''} This rigorous and methodological consistency effectively ensures that performance variations are attributable to architectural improvements rather than linguistic variance across evaluations.Geometric Pre-processing: Input images are automatically resized to spatial resolutions of 384 $\times$ 384 or 480 $\times$ 480. Corresponding hand keypoint annotations are extracted to compute the proposed OCAL supervision signals during the entire training phase.\\
\textbf{Geometric Pre-processing:} Input images are resized to $384 \times 384$ or $480 \times 480$. Corresponding hand keypoint annotations are extracted to compute the OCAL supervision signals during the training phase.

Prompt Engineering: 

%%%%%%%%%%%%%%%%%%%%%%%%%%%超参数表格1

\begin{table}[t]
\centering
\caption{Summary of Hyper-parameters in VistaRef.}
\label{tab:hyperparameters}
\resizebox{\linewidth}{!}{
\begin{tabular}{lll}
\toprule
\textbf{Category} & \textbf{Parameter} & \textbf{Value} \\
\midrule
\textbf{Architecture} & Embedding dimension ($d$) & 768 (Base) / 1024 (Large) \\
 & MLP hidden layers & $d$ \\
 & MLP layer count & 3 \\
 & Multi-head attention heads & 8 \\
 & Image input size & 384$\times$384 / 480$\times$480 \\
\midrule
\textbf{LHEM Module} & Center offset ratio & 0.3 \\
 & Long-axis scaling ratio & 2.2 \\
 & Short-axis scaling ratio & 1.0 \\
 & Minimum region size & $2.0 / \text{GridSize}$ \\
\midrule
\textbf{Feature Fusion} & Fixed gated scaling ($\gamma$) & 6.0 \\
\midrule
\textbf{Loss Weights} & Base loss weight ($\mathcal{L}_{base}$) & 0.7 \\
 & OCAL loss weight ($\mathcal{L}_{ocal}$) & 0.3 \\
 & Hand classification ($\lambda_{hand}$) & 0.2 \\
 & Keypoint regression ($\lambda_{kp}$) & 0.4 \\
 & Ray alignment ($\lambda_{ray}$) & 0.4 \\
\midrule
\textbf{Optimization} & Warm-up epochs & 10 \\
 & Fine-tuning epochs & 20 \\
 & EMA momentum & 0.99 \\
\bottomrule
\end{tabular}}
\end{table}

%%%%%%%%%%%%%%%%%% C

\section{Technical Remarks}

%%%%%%%%%%%%%%%%%%%%%%%%%%%%%%%%%%%%%%%%%%%%%%%%%%%%%%%%%%
\begin{figure*}
    \centering
    \includegraphics[width=\linewidth]{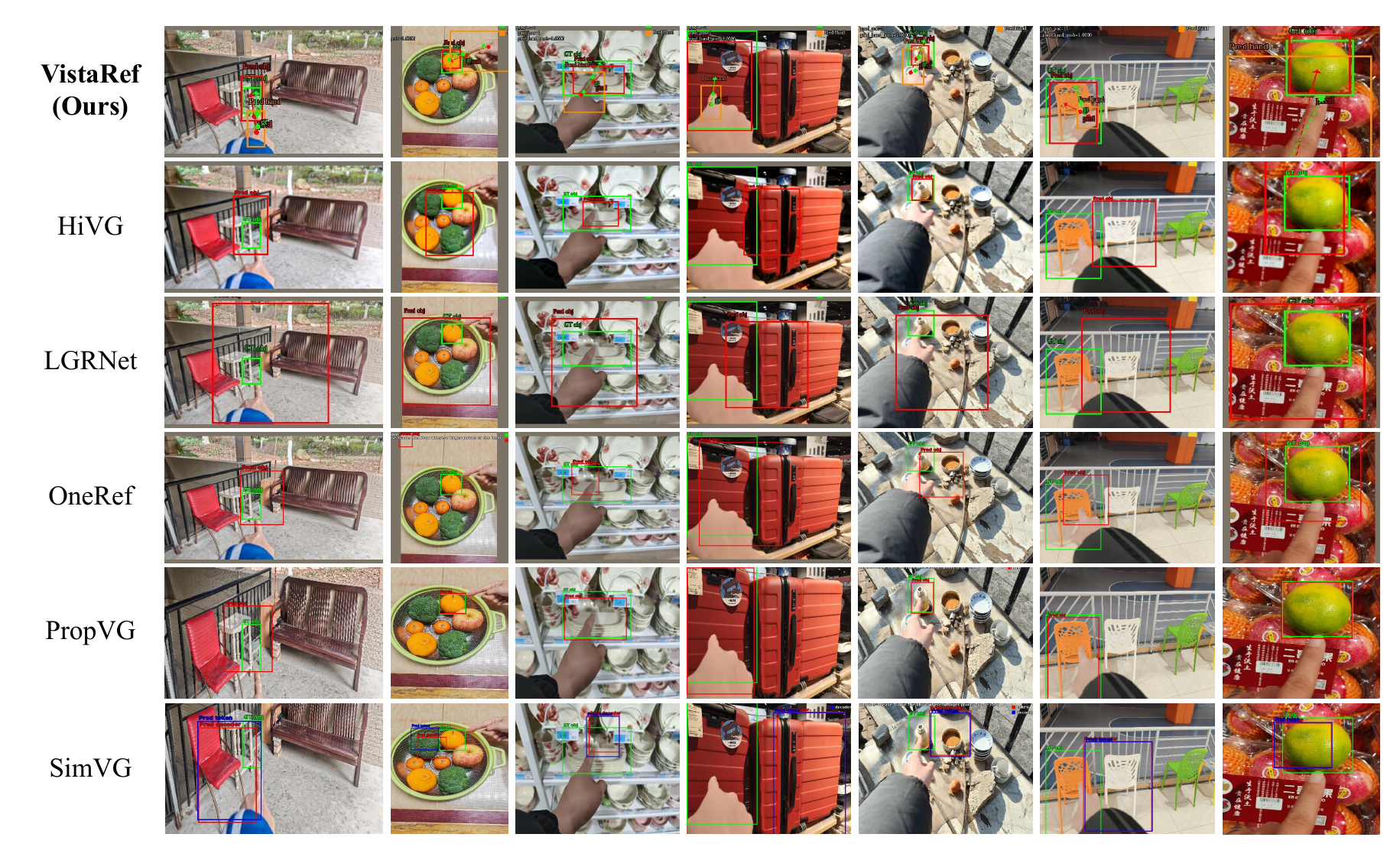}
    \caption{Comparison results in salient-object scenarios. Our model remains robust to salient distractors and accurately localizes the target indicated by the pointing gesture.}
    \label{fig:supple-result1}
\end{figure*}
%%%%%%%%%%%%%%%%%%%%%%%%%%%%%%%%%%%%%%%%%%%%%%%%%%%%%%%%

%%%%%%%%%%%%%%%%%%%%%%%%%%%%%%%%%%%%%%%%%%%%%%%%%%%%%%%%%%
\begin{figure*}
    \centering
    \includegraphics[width=\linewidth]{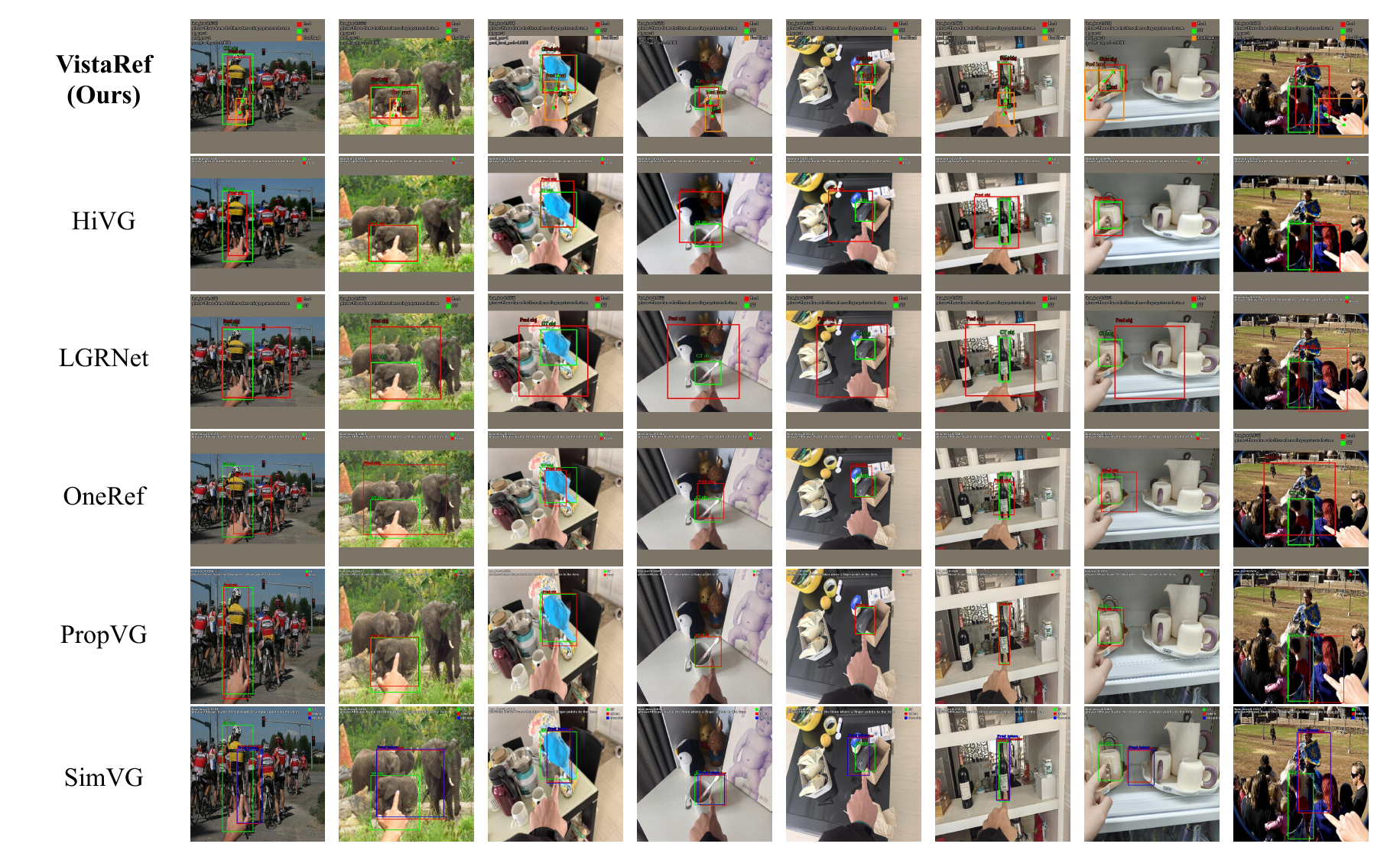}
    \caption{Comparison results in cluttered multi-object scenarios. Our model maintains reliable and accurate localization performance even in the presence of multiple objects and substantial visual interference.}
    \label{fig:supple-result2}
\end{figure*}
%%%%%%%%%%%%%%%%%%%%%%%%%%%%%%%%%%%%%%%%%%%%%%%%%%%%%%%%
%%%%%%%%%%%%%%%% C.1
\subsection{GRM Implementation}

\textbf{Keypoint Localization via $\mathbf{Head}_{kp}$:} The process of geometric modeling begins with the text-guided holistic visual feature $\mathbf{h}_{vl} \in \mathbb{R}^d$, which encapsulates the semantic alignment between the linguistic query and the global visual context. We feed this feature into a dedicated keypoint detection head, $\mathbf{Head}_{kp}$, comprising a three-layer MLP stack. This head regresses four raw logits that are subsequently transformed via a Sigmoid activation into normalized 2D coordinates within the unit interval $[0, 1]^2$. To ensure precise geometric consistency with the physical pointing gesture, we define these coordinates as the hand root $\mathbf{p}_{root}(r_x, r_y)$ and the fingertip $\mathbf{p}_{top}(t_x, t_y)$. Based on these localized points, we explicitly derive the directed pointing vector $\vec{\Delta} = \mathbf{p}_{top} - \mathbf{p}_{root}$, which serves as a geometric primitive representing the orientation and spatial extent of the pointing gesture. By conditioning the localization on $\mathbf{h}_{vl}$, the model inherently filters out irrelevant hand instances, ensuring that the predicted keypoints and the resulting vector $\vec{\Delta}$ strictly correspond to the referential intent of the user.\\
\textbf{Geometric Encoding and Ray Synthesis:} Following the derivation of the displacement magnitude $l = \|\mathbf{p}_{top} - \mathbf{p}_{root}\|_2$ and the unit direction vector $\vec{u}$, the module performs a latent projection to align these physical attributes with the Transformer's high-dimensional space. We first construct a 7-dimensional geometric primitive vector $\mathbf{g}_{ray}$ by explicitly concatenating the predicted coordinates and their derived attributes: $\mathbf{g}_{ray} = [\mathbf{p}_{root}(r_x, r_y); \mathbf{p}_{top}(t_x, t_y); \vec{u}; l]$. This vector is then processed by the Ray Encoder, a three-layer MLP that performs a non-linear mapping from the 7D physical space into the $d$-dimensional latent space, resulting in the ray embedding $\mathbf{e}_{ray} \in \mathbb{R}^d$. This encoding step is fundamental as it translates raw coordinate-based geometry into a high-dimensional feature format, enabling the subsequent cross-attention mechanism to utilize the geometric ray as a deterministic spatial prior for target grounding.

%%%%%%%%%%% C.2
\subsection{LHEM Implementation}
In the engineering implementation of the \textbf{LHEM} module, $\mathbf{\Omega}_{hand}$ is defined as a dynamic local hand region \textbf{adaptively} generated from the predicted keypoints to facilitate grounding.\\
\textbf{Region Construction:} For computational efficiency, the module calculates a logical rectangle aligned with the hand's principal axis and then extracts its \textit{axis-aligned bounding box} as the final $\mathbf{\Omega}_{hand}$.\\
\textbf{Geometric Priors:} The region center is fixed at $\mathbf{p}_{root} + 0.3 \cdot (\mathbf{p}_{top} - \mathbf{p}_{root})$. The long and short axes are dynamically scaled at 2.2$\times$ and 1.0$\times$ the ray length $l$, respectively. As shown in Fig. \ref{fig:supple-hand}, we present a more detailed process of fingertip coordinate localization.\\

\textbf{Robust Sampling and Boundary Handling:} To prevent sampling failure for small hand instances, we enforce a minimum region size of $2.0/GridSize$, ensuring $\mathbf{\Omega}_{hand}$ covers at least a $2 \times 2$ patch area. $GridSize$ represents the spatial resolution of visual tokens, calculated as the input image size divided by the patch size—resulting in a $GridSize$ of $24$ for $384 \times 384$ and $30$ for $480 \times 480$ resolutions. This spatial discretization ensures that the size constraint consistently encompasses a local patch neighborhood to facilitate robust gestural feature sampling and stable geometric alignment. If the sampling mask remains empty due to boundary conditions, a fallback mechanism selects the patch feature closest to the center.

%%%%%%%%%%%%%%% C.3
\subsection{Ray-guided Target Grounding}
\textbf{Cross-Attention Mechanism:} To capture the intricate spatial relationship between the pointing gesture and the potential target objects in the scene, we employ a multi-head cross-attention mechanism configured with 8 attention heads. In this architecture, the explicit gestural query $\mathbf{Q}_{rh} \in \mathbb{R}^d$ serves as the Query, which inherently carries the fused information of the local hand features and the geometric ray embedding. The global visual patch features $\mathbf{F}_v$ serve as the source of information, where the [CLS] token is specifically excluded to maintain a pure representation of the spatial environment. These patches are linearly projected to form the Key ($\mathbf{K}_{F_v}$) and Value ($\mathbf{V}_{F_v}$) matrices. By calculating the attention scores between the gestural query and all available image patches, the model is able to selectively attend to the most relevant visual regions that align with the pointing direction. This mechanism yields the interaction feature $\mathbf{t} \in \mathbb{R}^d$, which effectively encodes the refined cross-modal consensus between the user's physical gesture and the surrounding visual context.\\
\textbf{Synthesis of Integrated Representation $\mathbf{F}_{final}$:} In the final stage, the model constructs a comprehensive joint representation $\mathbf{F}_{final} \in \mathbb{R}^{3d}$ by directly concatenating three distinct $d$-dimensional feature vectors along the channel dimension: the text-guided visual feature $\mathbf{h}_{vl}$, the scaled ray embedding $\gamma \cdot \mathbf{e}_{ray}$, and the interaction features $\mathbf{t}$. This multi-source synthesis effectively expands the feature space to accommodate global semantic context, explicit geometric priors, and local gestural interactions within a single unified vector. We set the scaling factor $\gamma$ to $6.0$ to calibrate the numerical magnitude of the geometric ray prior, ensuring it is commensurate with the high-dimensional semantic features. This calibration prevents the deterministic geometric guidance from being overshadowed by the global context during the fusion process, thereby maintaining the structural integrity of the spatial reasoning chain.\\
\textbf{Target Localization via $\mathbf{MLP}_{bbox}$:} The integrated representation $\mathbf{F}_{final}$ serves as the input to the bounding box regression head $\mathbf{MLP}_{bbox}$, which decodes the fused multimodal information into explicit spatial parameters. This head consists of a three-layer MLP stack that performs a non-linear mapping from the $3d$-dimensional fused space to the 4-dimensional target space. The model regresses the target coordinates $\mathbf{b} = [x_{center}, y_{center}, w, h]\in [0, 1]^4$, which are subsequently normalized via a Sigmoid activation to the unit interval $[0, 1]$ relative to the image dimensions. By leveraging the geometric and semantic path encapsulated in $\mathbf{F}_{final}$, the regression head effectively translates the latent directional cues into a precise bounding area, enabling robust object grounding even in cluttered or distant scenarios.

%%%%%%%%%%%%%%%%%%% C.4

%%%%%%%%%%%TO DO by WXK%%%%%%%%%%%%%%%%%
%%增加一些模型细节%%%%

%%%%%

\section{Experimental Details}
We conduct experiments on the EgoPoint-Ground dataset, which is designed for egocentric pointing-based target localization. Unlike conventional visual grounding tasks, this setting emphasizes disambiguation via pointing gestures, in which the model is required to predict the bounding box of the object indicated by a pointing finger, given an input image and a fixed textual query. Each image is annotated with a hand instance, target objects, and finger keypoints, including the root and fingertip. During preprocessing, the hand annotation is identified and separated from the remaining object annotations, which are treated as candidate targets. For negative samples, the target box is set to zero and a binary label \texttt{is\_positive} is introduced. The textual query is fixed as \textit{``Please frame the bbox where a finger points to the item.''}  We adopt standard grounding metrics including Precision@0.3, Precision@0.5, Precision@0.7, and mIoU, where Precision@0.3 reflects coarse localization, Precision@0.5 is the primary metric, and Precision@0.7 evaluates high-precision localization. All results are reported on the test set.

Our method is built upon the OneRef framework with two backbone variants, BEiT-3 Base (\texttt{beit3\_base\_patch16\_384}) and BEiT-3 Large (\texttt{beit3\_large\_patch16\_384}), both initialized from MRefM pre-trained weights, and text inputs are tokenized using the official SentencePiece tokenizer. All input images are resized to $384 \times 384$ and the maximum query length is set to 64, with data augmentation including random cropping, scaling, and translation. During evaluation, the batch size is set to 128.

We adopt a standard single-dataset fine-tuning strategy with two training stages. In the warm-up stage, the backbone is frozen and the model is trained for 10 epochs with a batch size of 64 and a learning rate of $2.5\times10^{-4}$ using a cosine scheduler. In the fine-tuning stage, we load the best checkpoint from warm-up and continue training for 20 epochs with a batch size of 8 and a learning rate of $3\times10^{-5}$, while enabling box-mask constraints to improve localization accuracy.

The overall loss function follows the formulation described in Sec.~3.5. 
Specifically, it consists of the base detection loss $\mathcal{L}_{base}$ and the 
orientation-consistent alignment loss $\mathcal{L}_{ocal}$.In our experiments, the loss weights are empirically set as 
$\lambda_{base}=0.7$ and $\lambda_{ocal}=0.3$. 
Within $\mathcal{L}_{ocal}$, the component weights are configured as 
$\lambda_{hand}=0.2$, $\lambda_{kp}=0.4$, and $\lambda_{ray}=0.4$.
Following the asymmetric supervision mechanism described in Sec.~3.5, 
the keypoint loss $\mathcal{L}_{kp}$ and ray loss $\mathcal{L}_{ray}$ are 
only applied to positive samples with valid pointing gestures, while the 
hand classification loss $\mathcal{L}_{hand}$ is applied to all samples.

To stabilize optimization, we adopt an EMA-based loss normalization strategy with momentum $0.99$, $\epsilon=1\times10^{-6}$, and 100 warm-up iterations, which dynamically balances the contribution of different loss terms.

To evaluate the contributions of different components and supervision signals, we conduct ablation studies based on OneRef under a unified training setting, using BEiT-3 Base as the backbone. For core modules, we progressively introduce the GRM, LHEM, and the Cross-Attention mechanism into the baseline, while keeping training configurations unchanged, to analyze both individual contributions and their synergistic effects.

For ablation on individual loss terms, we study the impact of each supervisory signal in the OCAL framework by selectively enabling or disabling $\mathcal{L}_{hand}$, $\mathcal{L}_{kp}$, and $\mathcal{L}_{ray}$, while retaining the full model architecture. We adopt an incremental activation strategy, starting from the baseline with $\mathcal{L}_{base}$ only, and progressively incorporating additional loss terms. All settings share identical training configurations, allowing us to isolate the contribution of each loss and its combinations.

We further conduct hyperparameter sensitivity analysis using a one-factor-at-a-time strategy, where the remaining parameters are fixed to isolate the effect of each variable.

\section{Visualization Results}
\subsection{Qualitative Comparison}
As shown in Fig.~\ref{fig:supple-result1} and Fig.~\ref{fig:supple-result2}, our method consistently achieves more accurate and stable localization compared to existing approaches. 
In relatively simple scenarios (Fig.~\ref{fig:supple-result1}), baseline methods often exhibit noticeable deviations, such as bounding box drift toward salient but irrelevant regions or incomplete coverage of the target object. 
In contrast, our model produces tighter and more precise predictions aligned with the pointing direction.

In more challenging, cluttered multi-object scenarios (Fig.~\ref{fig:supple-result2}), the performance gap becomes more evident. 
Existing methods are more susceptible to visual distractions and tend to mislocalize toward nearby objects or larger salient regions. 
Benefiting from gesture-guided spatial constraints, our approach remains robust and can reliably identify the correct referent even under heavy clutter and ambiguity.
\subsection{Attention Map}
%%%%%%%%%%%%%%%%%%%%%%%%%%%%%%%%%%%%%%
%attention map
\begin{figure*}[!h]
    \centering
    \setlength{\tabcolsep}{1pt}
    \renewcommand{\arraystretch}{0}

    \begin{tabular}{ccccc}
        \includegraphics[width=0.19\linewidth]{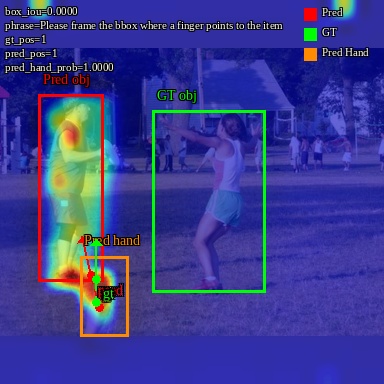} &
        \includegraphics[width=0.19\linewidth]{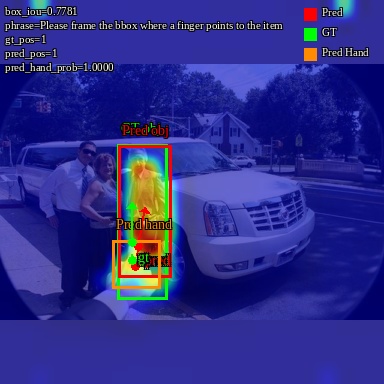} &
        \includegraphics[width=0.19\linewidth]{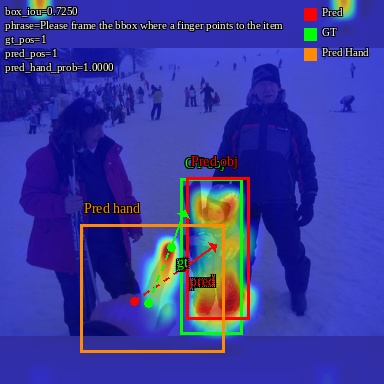} &
        \includegraphics[width=0.19\linewidth]{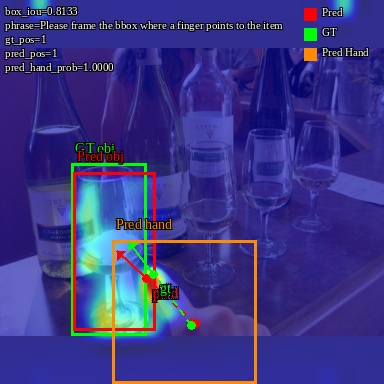} &
        \includegraphics[width=0.19\linewidth]{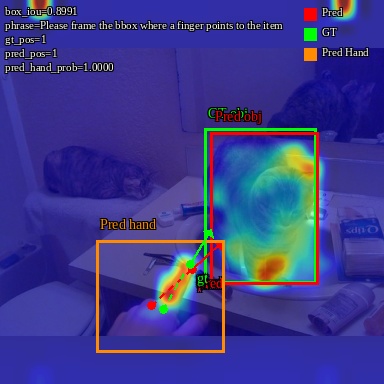} \\

        \includegraphics[width=0.19\linewidth]{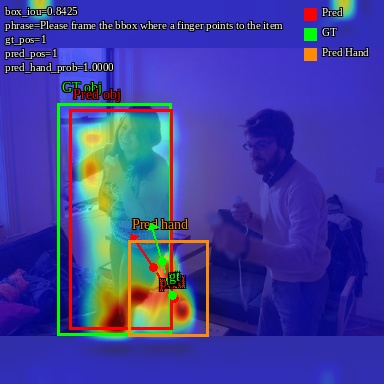} &
        \includegraphics[width=0.19\linewidth]{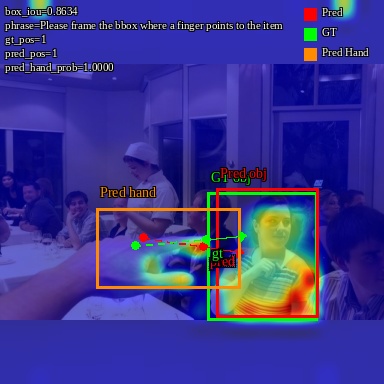} &
        \includegraphics[width=0.19\linewidth]{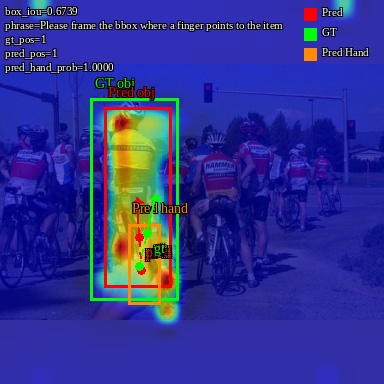} &
        \includegraphics[width=0.19\linewidth]{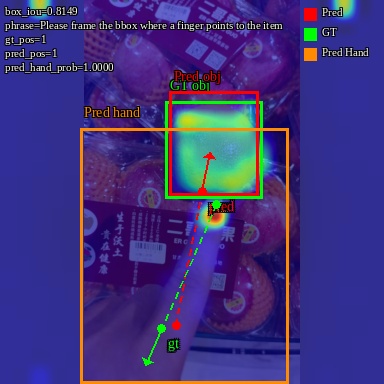} &
        \includegraphics[width=0.19\linewidth]{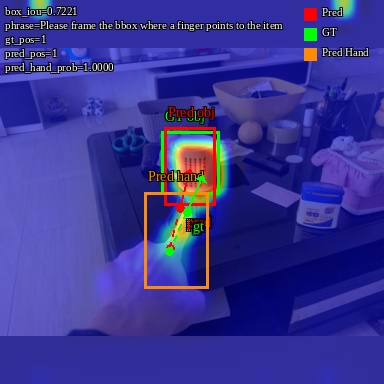} \\
    \end{tabular}

    \caption{
    Visualization of attention maps for representative samples.
    VistaRef produces spatially aligned and direction-aware attention distributions,
    concentrating along the pointing ray from the hand to the target object.
    Compared with conventional Transformer-based grounding models, the attention is more focused
    and less affected by distractors, demonstrating improved spatial reasoning ability.
    }
    \label{fig:attention_maps}
\end{figure*}
%%%%%%%%%%%%%%%%%%%%%%%%%%%%%%%%%%%%%%%%%%%%%%%%%%%%%%%%%%%%%%%%%%%%
\begin{figure*}
    \centering
    \includegraphics[width=\linewidth]{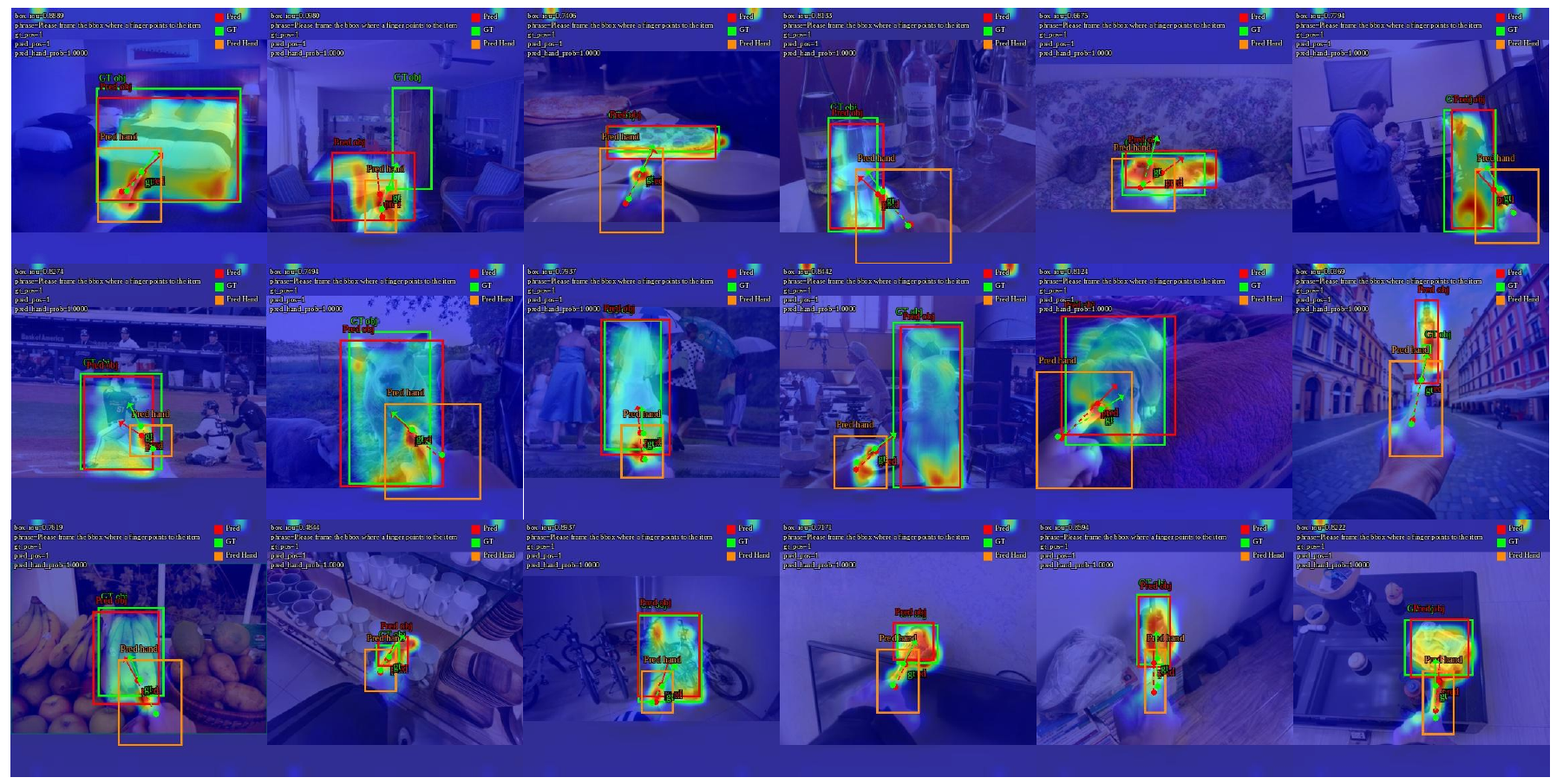}
    \caption{ More Visualization of attention maps for representative samples.}
    \label{fig:placeholder}
\end{figure*}

To further understand how VistaRef enhances spatial orientation awareness, we visualize attention maps from representative samples in Fig.~\ref{fig:attention_maps}. These visualizations provide direct evidence that the proposed geometric modeling reshapes the attention distribution compared with conventional Transformer-based grounding models.

\medskip

\noindent\textbf{Alignment with Geometric Ray.}
A key observation is that the attention distribution in VistaRef exhibits a clear anisotropic pattern aligned with the pointing direction. Instead of spreading uniformly across semantically relevant regions, the attention progressively concentrates along the implicit geometric ray from the hand root to the fingertip, and further extends toward the target object.

This behavior directly reflects the effect of the Geometric Ray Modeling (GRM) module, which transforms the implicit pointing cue into an explicit directional prior. By encoding the ray as a structured feature and injecting it into the attention mechanism, the model performs direction-aware feature aggregation rather than purely semantic matching.

\medskip

\noindent\textbf{Suppression of Irrelevant Regions.}
Compared with baseline attention patterns, which often highlight multiple semantically similar objects, VistaRef shows significantly reduced activation on distractors, even when they share similar appearance or category with the target. This behavior can be attributed to the combined effect of two mechanisms: the ray constraint ensures that only regions consistent with the pointing direction receive high attention weights, while the ray-aware cross-attention selectively retrieves spatially aligned features from the global visual representation. As a result, the model effectively suppresses irrelevant regions and avoids the common failure mode of semantic ambiguity, where multiple objects compete for attention.

\medskip

\noindent\textbf{Localized Hand-Centric Attention Initialization.}
The attention maps consistently show strong activation around the hand region, especially near the fingertip. This demonstrates the effectiveness of the Local Hand Entity Modeling (LHEM) module, which extracts a purified gesture representation from a dynamically constructed local region $\Omega_{\text{hand}}$.

Instead of relying on global features, the model first anchors attention to the gesture origin and then propagates it along the ray. This ``hand $\rightarrow$ ray $\rightarrow$ target'' progression matches the physical pointing process and forms a causal attention chain.

\medskip

\noindent\textbf{Progressive Attention Transition (Hand $\rightarrow$ Target).}
Across multiple samples, we observe a consistent attention transition pattern, where the attention initially concentrates around the hand region, then gradually elongates along the pointing direction, and finally aggregates on the target object. This structured transition indicates that the model no longer treats grounding as a static matching problem, but instead performs a geometric reasoning process over spatial trajectories. Such behavior is absent in conventional Transformer attention, which lacks explicit mechanisms to model directional continuity.

\medskip

\noindent\textbf{Robustness in Cluttered and Long-Range Scenarios.}
In challenging cases involving densely packed objects, small targets, or long pointing distances, the attention maps of VistaRef remain highly directional and stable, without diffusing into irrelevant regions.

This observation confirms that explicit geometric modeling helps mitigate orientation insensitivity and localization drift, particularly in complex real-world scenes.

% \appendix









\newpage
%%%%%%%%%%%%%%%%%%%%%%%%%  参考文献列表
\bibliographystyle{ACM-Reference-Format}
\bibliography{sample-base}

\end{document}